\documentclass[10pt,twocolumn,letterpaper]{article}

\usepackage{iccv}
\usepackage{times}
\usepackage{epsfig}
\usepackage{graphicx}
\usepackage{amsmath}
\usepackage{amssymb}
\usepackage{booktabs}
\usepackage{multirow}
\usepackage{authblk}
\usepackage{flushend}
\usepackage{animate}
\usepackage{subfigure}

\usepackage[pagebackref=true,breaklinks=true,letterpaper=true,colorlinks,bookmarks=false]{hyperref}

\iccvfinalcopy 

\def\iccvPaperID{5915} 

\ificcvfinal\fi

\begin{document}

\title{Unpaired Multi-domain Attribute Translation of 3D Facial Shapes with a Square and Symmetric Geometric Map}

\author[1,2]{Zhenfeng Fan}
\author[1,2]{Zhiheng Zhang}
\author[1,2]{Shuang Yang}
\author[1,2]{Chongyang Zhong}
\author[3]{Min Cao}
\author[1,2\thanks{The corresponding author}]{Shihong Xia}

\affil[1]{Institute of Computing Technology, Chinese Academy of Sciences}
\affil[2]{University of Chinese Academy of Sciences}
\affil[3]{Soochow University}

\affil[ ]{\tt\small {\{fanzhenfeng,zhangzhiheng20g,yangshuang21b,zhongchongyang,xsh\}@ict.ac.cn; mcao@suda.edu.cn}}


\maketitle
\ificcvfinal\thispagestyle{empty}\fi

\begin{abstract}
    While impressive progress has recently been made in image-oriented facial attribute translation, shape-oriented 3D facial attribute translation remains an unsolved issue. This is primarily limited by the lack of 3D generative models and ineffective usage of 3D facial data. We propose a learning framework for 3D facial attribute translation to relieve these limitations. Firstly, we customize a novel geometric map for 3D shape representation and embed it in an end-to-end generative adversarial network. The geometric map represents 3D shapes symmetrically on a square image grid, while preserving the neighboring relationship of 3D vertices in a local least-square sense. This enables effective learning for the latent representation of data with different attributes. Secondly, we employ a unified and unpaired learning framework for multi-domain attribute translation. It not only makes effective usage of data correlation from multiple domains, but also mitigates the constraint for hardly accessible paired data. Finally, we propose a hierarchical architecture for the discriminator to guarantee robust results against both global and local artifacts. We conduct extensive experiments to demonstrate the advantage of the proposed framework over the state-of-the-art in generating high-fidelity facial shapes. Given an input 3D facial shape, the proposed framework is able to synthesize novel shapes of different attributes, which covers some downstream applications, such as expression transfer, gender translation, and aging. Code at \href{https://github.com/NaughtyZZ/3D_facial_shape_attribute_translation_ssgmap}{https://github.com/NaughtyZZ/3D\_facial\_shape\_attribute\_tr\\anslation\_ssgmap}.
\end{abstract}

\section{Introduction}
\label{sec:intro}
\begin{figure}[htbp]
	\begin{center}
		\includegraphics[width=1\linewidth]{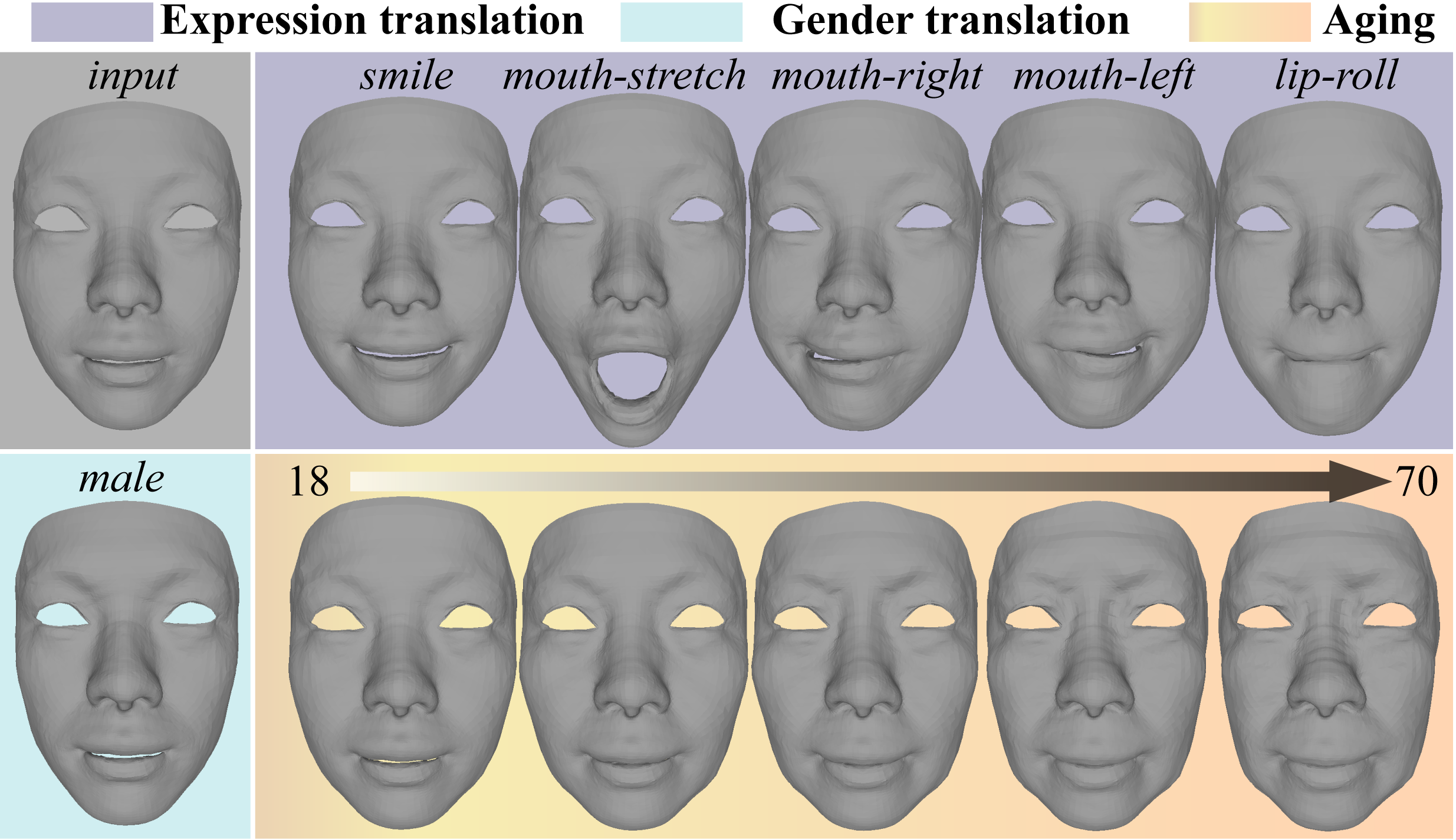}
	\end{center}
	\caption{Translation of 3D shape attributes to multiple domains that correspond to expressions, genders, and ages.}
	\label{fig_show}
\end{figure} 

The advancement of generative adversarial networks (GANs) has recently activated a lot of studies on face-related tasks, such as face synthesis~\cite{karras2019style,bao2021high}, facial image super-resolution~\cite{ledig2017photo,isola2017image}, and facial attribute translation~\cite{choi2018stargan,abdal2021styleflow}. The rationale of these studies is to use an auxiliary discriminator network to regularize the output face with the learned distribution of facial data towards a specific category. The task of facial attribute translation is referred to as changing the particular aspect of a facial image, \textit{e.g.} changing the expression~\cite{fan2019controllable} or the age~\cite{yang2018learning} of a face, resulting in desired appearances. This task has many downstream applications in the media and film industry.

In the past, most methods in the field of facial attribute translation dealt with 2D facial images~\cite{chang2021learning,pumarola2018ganimation,shen2020interfacegan}. This is partly because 2D facial images are easily accessible and partly because deep learning based methods commonly work on image data with a regular grid. Nowadays, researchers are paying more attention to 3D faces~\cite{egger20203d,tewari2021learning,ploumpis2020towards,egger2018occlusion,gafni2021dynamic,zheng2022avatar,park2019deepsdf} for wider applications and more realistic rendering, due to advances in 3D imaging sensors and 3D applications. Editing 3D facial shape has attracted much interest in both the computer graphics and computer vision communities. Translation of 3D facial shapes provides geometric flexibility in addition to textures, thus resulting in vividness for facial rigging and animation~\cite{geng2018warp,li2010example,richard2021meshtalk,grassal2022neural}. The 3D shape is also considered as a vital commodity of face to overcome pose and illumination challenges in existing face recognition literature~\cite{liu2018joint,zhao20183d,liu2018disentangling}. 

In this work, we study the problem of 3D facial shape attribute translation. An example is shown in Figure~\ref{fig_show}.
We denote \emph{attributes} as inherent features of a 3D face, such as \textbf{expression}, \textbf{age}, and \textbf{gender} that correlate to shape variations. We also refer to some GAN-based domain adaptation works~\cite{choi2018stargan, isobe2021multi}, and define ``\emph{domain}'' as a set of data with certain attribute values, \textit{e.g.} $30$-year-old males with neutral expression.  We cast 3D facial attribute translation as a domain adaptation problem, which is naturally linked with GAN by its data-driven nature. 

Applying state-of-the-art deep GANs on 3D geometry data is challenging, and the difficulties are mainly two-fold.  Firstly, unlike facial images on a square Euclidean grid, 3D geometric data, in the case of facial surfaces, are on a Riemannian manifold. This hinders the application of state-of-the-art deep convolutional neural networks (CNNs) on 3D facial attribute translation. We refer to this as a problem of \emph{network compatibility}. Secondly, unlike facial image data that are abundant for image translation tasks, there is a shortage of 3D facial data. This is limited by the popularity of 3D scanning devices. Moreover, most deep CNN based methods rely heavily on paired and labeled data. We refer to this as a problem of \emph{data scarcity}.

For \emph{network compatibility}, we design a geometric map that encodes 3D coordinates onto regular image grids. The adjacency information for 3D vertices is preserved in a \emph{local least-square} manner while being constrained by \emph{symmetric} property. This enables us to leverage symmetry, an important character of face in the learning process. In addition, the learning networks are \emph{end-to-end} with a differentiable 3D-to-2D forward geometric mapping layer and a 2D-to-3D backward grid sampling layer.

For \emph{data scarcity}, we employ a \emph{unified} and \emph{unpaired} GAN for multi-domain attribute translation. Firstly, we assume that the latent encodings of different domains should be cross-correlated to each other. Rather than learning the translation between every two domains separately, we learn a single generator for all domains, \textit{i.e.} for expression, age, and gender translation tasks together. Secondly, we employ an unpaired framework, assuming that exactly paired data, \textit{i.e.} different ages of a person are difficult to collect and different genders for the same identity are almost impossible in the real world. This mitigates the exact constraint for hardly accessible paired data. We also conduct data augmentation in training by adding random perturbations of scales and rotations of 3D facial shapes. 

In summary, the main contributions of this paper are:
\begin{itemize}
	\item We first propose a general and unified framework for multi-domain 3D facial attribute translation, which covers some shape-oriented applications including expression transfer, aging, and gender translation.
	\item We construct a novel geometric map for 3D face representation on a canonical 2D grid. The geometric map leverages symmetry of face and maintains the adjacency of 3D vertices in a local least-square manner.
	\item We make unpaired training of 3D facial shape data available on a geometric map with a hierarchical GAN architecture to suppress both global and local artifacts.
\end{itemize}

\section{Related Work}
The closely related fields to this work include GAN-based 2D image translation and 3D face manipulation with UV maps. We now briefly discuss the most related works in each field, respectively.

\subsection{GAN-based 2D Image Translation}
GAN is very popular for generating novel and realistic images~\cite{isola2017image,liu2017unsupervised,zhan2021unbalanced,zheng2021spatially} because of its capability for modeling the distribution of a large amount of data. Many variants of GANs with different characteristics are proposed after the seminal work of Vanilla GAN~\cite{goodfellow2014generative}. 

Conditional GANs (CGANs)~\cite{mirza2014conditional,odena2017conditional} are originally proposed to generate samples conditioned on a specific class. CGANs commonly include class labels in both the generator and discriminator that are relevant to a specified task. For example, pix2pix~\cite{isola2017image} learns image-to-image translation with a CGAN architecture in a supervisory manner with paired data; Attgan~\cite{he2019attgan} is able to change the specified attribute of a facial image.

CycleGAN~\cite{zhu2017unpaired} and DisCoGAN~\cite{kim2017learning} release the paired data assumption by incorporating a cycle consistency loss that preserves the key attribute shared by input and output images. However, these GANs are limited to attribute translation tasks between two domains. Thus they cannot effectively make use of data that cover multiple domains. To alleviates the deterministic mapping problem, Huang \textit{et al.} propose an MUNIT framework~\cite{huang2018multimodal} by incorporating a style code to generate versatile images.

StarGANs~\cite{choi2018stargan,choi2020stargan} can learn multi-domain attribute translation with the help of domain classification loss in addition to cycle consistency loss and adversarial loss. Generally, they can effectively learn both the global features shared by data in all domains and the local features hold by data in a specific domain. U-GAT-IT~\cite{KimKKL20} and Attentiongan~\cite{TangLXTS23} further employ the attention mechanism to generate high-quality foreground against background.

In this work, we borrow some key architectures in StarGANs~\cite{choi2018stargan,choi2020stargan} for 2D images and design a geometric map elaborately for 3D facial data. We further incorporate them in an end-to-end adversarial learning framework with a novel hierarchical discriminator. Therefore, the proposed framework supports multi-domain 3D facial attribute translation in an unpaired manner with a single generator. 

\subsection{UV Maps for 3D Face Manipulation}
In a departure from 2D images with pixels on a regular grid, raw 3D geometric data are commonly organized as irregular points. A common way to represent 3D facial data is to flatten the 3D surface onto a 2D UV plane, on which the locations of 3D vertices are encoded. 

Blanz and Vetter~\cite{blanz1999morphable} project the facial surface onto a cylindrical UV map for shape registration in their seminal work for 3D Morphable Model (3DMM). The 3D location of each vertex is encoded as height and distance to the cylindrical axis. The 3DMM is further used for an attribute translation task to manipulate weight, gender, and expressions~\cite{amberg2009weight}. 

Bagautdinov \textit{et al.}~\cite{bagautdinov2018modeling} conduct non-rigid registration of 3D facial scans and deform a template face (as the average of all registered results) to a 2D plane. They further define a $3$-channel image (UV map) surrounding the deformed template. This ensures topological neighbors on 3D are also topological neighbors on the image. This strategy to represent 3D facial data is used by many recent works~\cite{abrevaya2019decoupled,gecer2020synthesizing,ma2021pixel,Stylianos20203DFaceGAN} for 3D faces. Generally, the locations of 3D vertices can be converted into an image-like tensor to apply the 2D convolutions~\cite{gecer2020synthesizing}. An advantage over some other works with voxel/point/graph convolution~\cite{jackson2017large,ranjan2018generating,cheng2019meshgan,lin2020towards} is the high-frequency details can be preserved better.

In this work, we construct a novel UV map that supports forward and backward differentiable operations to be embedded in an end-to-end learning network. We call our UV map a geometric map because it has the following novel geometric properties: 1) It is square; 2) It is symmetric with respect to a central axis; 3) the mapping between 3D vertices and their 2D correspondences is in a local least-square manner to preserve the 3D adjacency information as rigidly as possible. These properties enable us to learn a 3D attribute translation task effectively.

\section{Method}
\begin{figure*}[htbp]
	\begin{center}
		\includegraphics[width=1\linewidth]{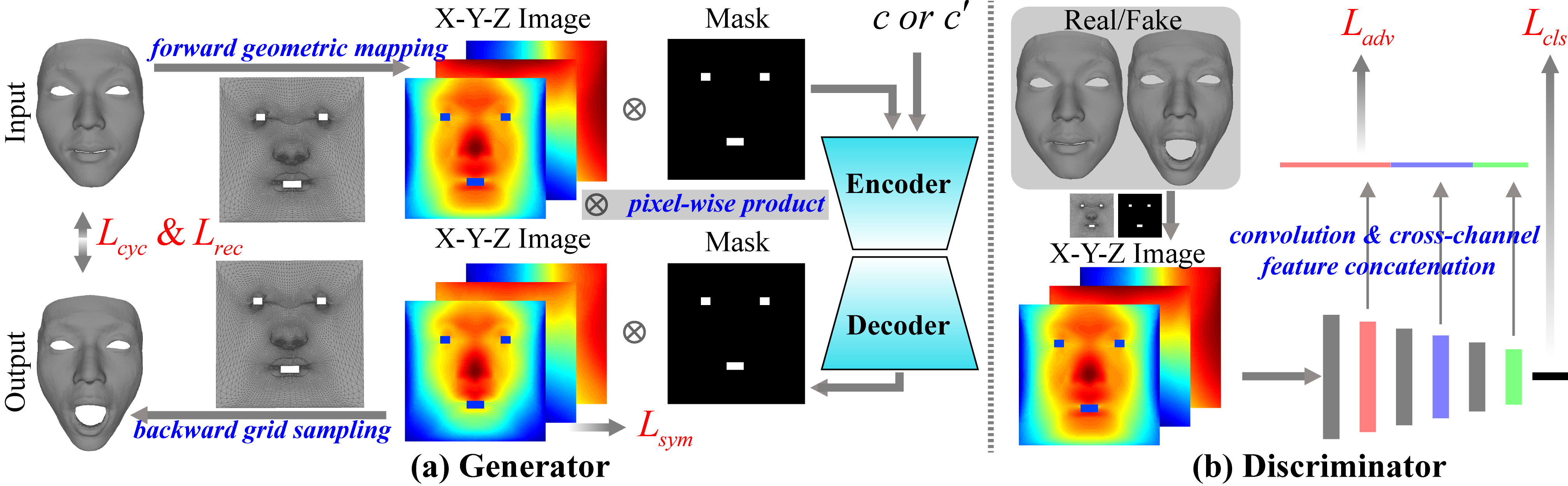}
	\end{center}
	\caption{The overall network architecture with the proposed geometric map. The loss functions for training are marked in red typeface.}
	\label{fig_network}
\end{figure*}

In this section, we describe the proposed geometric map, the network architecture, and the detailed loss functions in training, respectively, which constitute the basic elements of the proposed framework (see Figure~\ref{fig_network}). 

\subsection{Geometric Map Construction}
We consider triangle facial meshes as the input and output of the proposed method with the same number of vertices and the same mesh topology. Generalization to other formats of data, \textit{e.g.} point cloud, is applicable by rigidly alignment~\cite{paul1992registration,yang2015go} and non-rigid registration~\cite{amberg2007optimal,booth2018large} to a common template mesh in our implementation. Since all faces share the same topology, we denote each face by $\mathcal{V}\in\mathbb{R}^{3\times n}$ for simplicity. In order to make 2D deep CNN architectures compatible with 3D facial shapes, we construct a geometric map that bridges the gap between a 3D surface and a 2D image grid. The steps are: 

I. Average over some registered facial meshes to acquire a noiseless 3D template mesh $\mathcal{V}^{s}$;

II. Initialize the geometric map by harmonic parametrization~\cite{eck1995multiresolution,gu2002geometry} of $\mathcal{V}^{s}$ to $\mathcal{V}^{t}$;

\begin{figure}[htb]
	\begin{center}
		\includegraphics[width=1\linewidth]{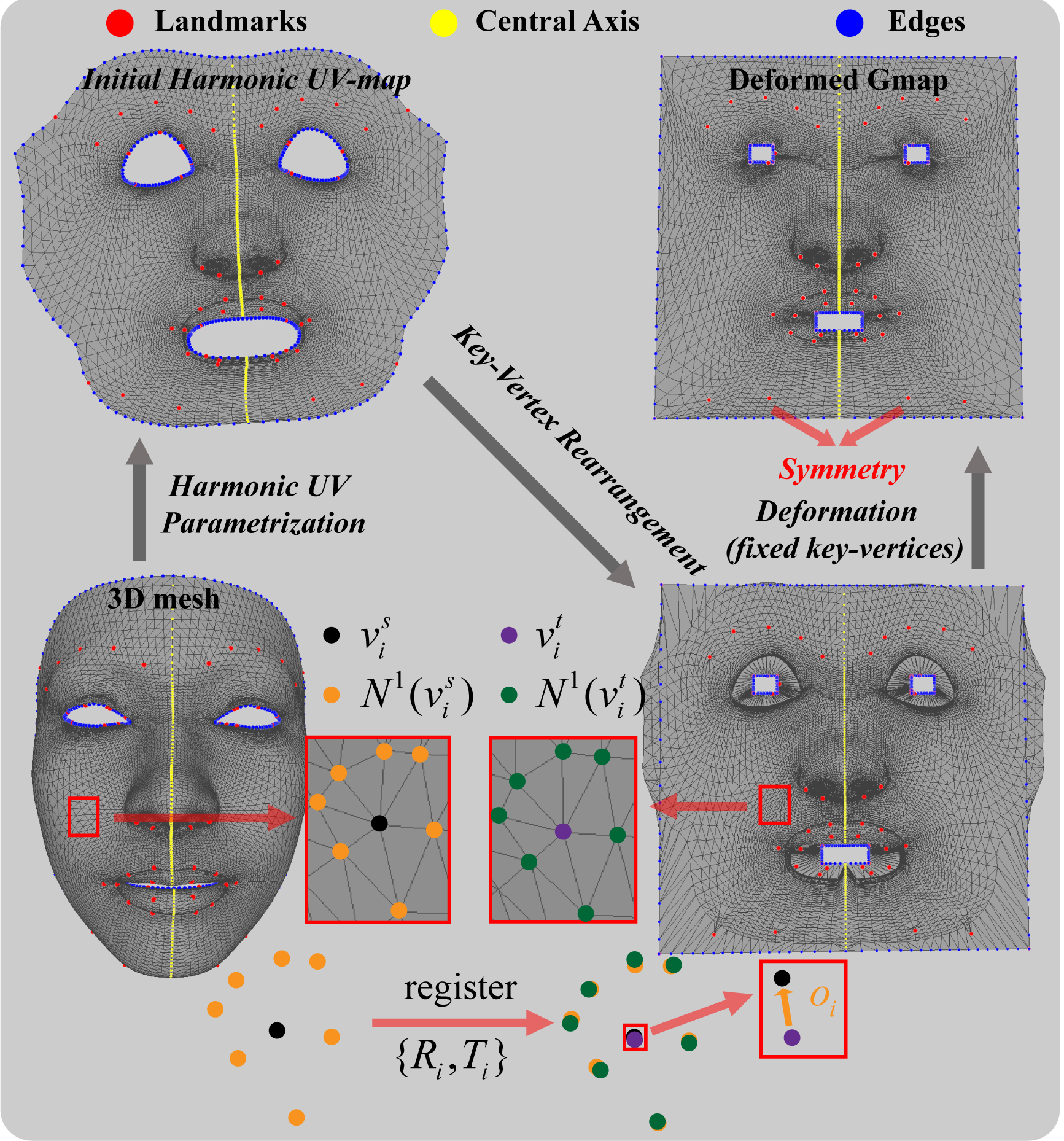}
	\end{center}
	\caption{The process to deform an initial harmonic UV-map to a square and symmetric geometric map (\textbf{Gmap}).} 
	\label{fig_gmap2}
\end{figure}

III. Deform $\mathcal{V}^{t}$ to a \textbf{square and symmetric} geometric map guided by the local structure of $\mathcal{V}^{s}$ and some rearranged key vertices, as follows (also refer to Fig.~\ref{fig_gmap2}).

First, we select some key vertices on $\mathcal{V}^{t}$ that are landmarks, edges, and central axis on the face. We adjust the central vertices to form the central axis. The locations of the landmarks and edges are rearranged to be square and symmetric with respect to the central axis.

Then, we denote the $1$-ring neighbors of each vertices $v_i^s \in \mathcal V^s$ as ${\mathcal N}^1(v_i)$. The correspondence of each vertex in $\mathcal{V}_s$ to $\mathcal{V}_t$ is determined by the subscript $i$. We suppose there exists a rigid transformation $\{R_i,T_i\}$ that aligns $v_i^s$ to $v_i^t$, as
\begin{equation}\label{e_add1}
{v_i^t} \leftarrow {R_i}{v_i^s} + {T_i}(i \in \mathcal V^s), 
\end{equation} 
where $\{R_i,T_i\}$ can be estimated by a least-square alignment problem of the surrounding $1$-ring neighbors, as 
\begin{equation}\label{e_add2}
\{ {R_i},{T_i}\}  = \mathop {\arg \min}\limits_{{R_i \in SO(3)},{T_i \in \mathbb R^3}} \sum\limits_{v_j^s \in {\mathcal N}^1(v_i)} {\left\| {{R_i}v_j^s + {T_i} - v_j^t} \right\|_2^2}.
\end{equation} 
Here $SO(3)$ denotes the space of all Givens matrices. 

Next, the preliminary offset for each vertex is regularized by local smoothness. We denote $p_i^t={R_i}{v_i^s} + {T_i}$ as the unregularized offset and formulate the problem as
\begin{equation}\label{e_add3}
\begin{aligned}
\begin{split}
\{ {o_i}|i \in \mathcal V^t\}  &= \mathop {\arg \min }\limits_{\{ {o_i}|i \in \mathcal V^t\} } \{ \sum\limits_{i \in \mathcal V^t} {\left\| {{p_i^t} - ({v_i^t} + o_i)} \right\|_2^2}\\
&+  \sum\limits_{i \in \mathcal V^t} {\sum\limits_{j \in {\mathcal N}^1(v_i)} {\left\| {{o_i} - {o_j}} \right\|_2^2} \} }.
\end{split}
\end{aligned}
\end{equation} 

Solving Eq.~\ref{e_add3} requires taking the partial derivative with respect to each offset $o_i(i \in \mathcal V^t)$ and leads to a linear system
\begin{equation}\label{e_add4}
[{{\bf{A}}_{ij}}]_{n \times n} \cdot {[{{\bf{O}}_{ij}}]_{n \times 3}} = {[{{\bf{B}}_{ij}}]_{n \times 3}},
\end{equation}
where
\begin{equation}\label{e_add5}
{{\bf{A}}_{ij}} = \left\{ {\begin{array}{*{20}{c}}
	{\noindent {1+2N_{v_i}}\quad\,\, \mbox{if}\quad i = j \,\,\qquad\qquad\qquad\,\,\,\,\,\,\,\,\,\,}  \\
	{\noindent  -2 \qquad\,\,\,\,\,\,\,\, \,         \mbox{if}\,\,\,\,\,\,i \ne j \,\, \mbox{and} \, j \in {\mathcal N}^1(v_i)} \quad \\
	{\noindent 0 \quad\quad\,\, \quad\quad \mbox{otherwise}\,\,\,\,\,\quad\,\,\,\, \quad\quad\quad\quad\quad}  \\
	\end{array}} \right.,
\end{equation}
\begin{equation}\label{e_add6}
{[{{\bf{O}}_{ij}}]_{n \times 3}} = {\left[ {{o}_1,...,{o}_n} \right]^T},
\end{equation}
and
\begin{equation}\label{e_add7}
{[{{\bf{B}}_{ij}}]_{N \times 3}}  = [p_1^t-v_1^t,...,p_n^t-v_n^t]^T.
\end{equation} 
$N_{v_i}$ is the number of vertices in ${\mathcal N}^1(v_i)$ and the superscript $T$ denotes matrix transpose.

Since some key vertices on the \textbf{landmarks, edges, and central axis} are fixed as constraints for solving Eq.~\ref{e_add3}, we exclude the corresponding columns in $\bf{A}$ and rows in $\bf{O}$. Therefore, $\bf{A}$ is a \emph{rank-deficient} matrix. Let $n_f$ be the number of fixed vertices ($\mathcal V^f \subset \mathcal V^t$), then Eq.~\ref{e_add4} is degraded to
\begin{equation}\label{e_add8}
{[{{\bf{A}}_{ij}}]_{n \times (n-n_f)}} \cdot {[{{\bf{O}}_{ij}}]_{(n-n_f) \times 3}} = {[{{\bf{B}}_{ij}}]_{(n-n_f) \times 3}},
\end{equation}
which is an \emph{over-determined} linear system. Its \textit{least-square} solution is given by the \textit{Moore-Penrose inverse} as
\begin{equation}\label{e_add9}
{\bf{O}}=({{\bf{A}}^T {\bf{A}}})^{-1}{\bf{A}}^T {\bf{B}}.
\end{equation}
By this way, we are able to fix some key vertices while updating other vertices. The vertex on the target mesh is added by each offset in $\bf{O}$ after solving Eq.~\ref{e_add8}, as
\begin{equation}\label{e_add10}
v_i^t=v_i^t+o_i(i\in \mathcal V^t/\mathcal V^f).
\end{equation} 

Finally, the steps from Eq.~\ref{e_add1} to Eq.~\ref{e_add10} are iterated until the ensembled offset is smaller than a certain threshold.

The above process is in fact a variant of a locally rigid registration process in~\cite{fan2023towards}. It conducts as rigid as possible alignments from the local cells ($1$-rings) on the original 3D mesh to the deformed geometric map, while being constrained by some rearranged key vertices. It also demonstrates that the proposed geometric map preserves the adjacency relationship\footnote{This brings about an advantage over the existing UV maps~\cite{bagautdinov2018modeling,abrevaya2019decoupled,wu2021adversarial}. The vertices on the upper and lower mouth area are separated, avoiding interferences from the local convolutional kernels in common CNNs.} of all vertices in a \textit{least-square} manner. 
This makes the mapping from 3D mesh to 2D geometric map to be one-to-one for each vertex, avoiding triangle flipping which is correlated to interpolation errors. Furthermore, as described in~\cite{gu2002geometry,sheffer2007mesh}, sampling the locations in a geometric image is prone to large interpolation errors unless the border vertices are preassigned to distinct pixels. In this work, there is seldom triangle flips even on the border regions, thus avoiding most interpolation errors.

In addition, the advantages for the geometric map to be \textit{square and symmetric} are: 1) making full use of the pixels (2D images also cover a square area); 2) easy to exclude the blank pixels inside the eyes and mouth; 3) easy to leverage a symmetric loss in training by image flip operation.

\subsection{Network Architecture}

The network architecture\footnote{The detailed architecture  is described in the supplementary material.} employs a main generator network and an auxiliary discriminator network, as in Figure~\ref{fig_network}. The mapping between 3D vertices and 2D image grid is computed by \emph{forward barycentric interpolation} (as in~\cite{abrevaya2019decoupled}) and \emph{backward bilinear grid sampling} with the aforementioned geometric map\footnote{The detailed exposition for the forward and backward mappings is described in the supplementary material.}. A mask for the valid pixels is customized accordingly. In a departure from some existing works, the forward and backward mappings are incorporated into our network in an end-to-end manner. During training, the generator network transfers the 3D facial shape to a certain domain, while the discriminator network enforces indistinguishable generation (probability distribution) to the given domain.  

\textbf{Generator.} We use an encoder-decoder architecture for the generator network. The feature expansion of the bottleneck is only downsampled by a factor of $4$ to avoid the mixing of spatial content (thus friendly to high-frequency information) of the input geometric map. We also include $6$ residual blocks~\cite{he2016deep} in the bottleneck. The skip connection of the residual block is able to learn the latent representations of data in multiple domains efficiently. Moreover, we embed both the forward geometric mapping layer and backward grid sampling layer as the input and output layers of the network, respectively. The end-to-end learning setting can compensate for the sampling errors between a 3D face and its representation on the geometric map.

\textbf{Discriminator.}  Previous studies discriminate the feature in different dimensions, \textit{e.g.} Vanilla GAN~\cite{goodfellow2014generative} for the global image, Pixel-GAN~\cite{isola2017image} for each pixel, and Patch-GAN~\cite{li2016precomputed} for a few intersected receptive fields on an image. In a departure from that, we employ a strategy that we call Pyramid-GAN to discriminate both global and local patterns of the input shape. The discriminator downsamples the feature expansion by $2$ in a cascade manner until the height and width are $2 \times 2$. Specifically, we conduct bifurcated convolutional operations on every other feature layer, then flatten and concatenate each output, and finally fed them to the adversarial loss. This strategy leads to both globally and locally realistic generations of 3D faces.

\subsection{Loss Function} 

We employ several loss functions (also refer to Figure~\ref{fig_network} for the exact locations) in the training process. The symbols of some variables are listed below for brevity. 

\begin{itemize}
	\setlength{\itemsep}{0pt}
	\setlength{\parsep}{0pt}
	\setlength{\parskip}{0pt}
	\item $x$ denotes the input shape of the generator fed into the forward geometric mapping layer.
	\item $y$ denotes the output geometric map of the generator fed into the backward grid sampling layer.
	\item $c$ and $c'$ are the target and source domain labels of $x$, respectively, which are composed of discrete expression and gender labels and continuous age labels.
	\item $G$ denotes the generator network; $D_{src}$ and $D_{cls}$ denotes the real/fake and domain classification branch of the discriminator network, respectively.
\end{itemize}

\textbf{The Adversarial Loss} forces the distribution of the generated
3D shapes to approach the shapes of real faces, as
\begin{equation}\label{e3}
L_{adv} = E_x[\log D_{src}(x)] + E_{x,c}[\log (1-D_{src}((G(x,c))],
\end{equation}
given the generator taking as input 3D shape $x$ conditioned on domain label $c$. In practice, we adopt the Wasserstein GAN~\cite{arjovsky2017wasserstein} with gradient penalty to stabilize the training process as
\begin{equation}\label{e4}
\begin{aligned}
L_{adv} = &E_x[\log D_{src}(x)] - E_{x,c}[D_{src}((G(x,c)\\
&-\lambda_{gp}E_{\hat x}[({\left\| {{\nabla _{\hat x}}D_{src}(\hat x)} \right\|_2} - \alpha)^2]],
\end{aligned}
\end{equation}
where $\hat x$ is uniformly sampled between a pair of real and generated shapes. And we set $\alpha$ to $0.01$.

\textbf{The Classification loss} enables effective domain translation from $c'$ to $c$. The objective is decomposed
into
\begin{equation}\label{e5}
L_{cls}^r = E_{x,c'}[-\log D_{cls}(c'| x)]
\end{equation}
for real shapes to optimize the discriminator, and  
\begin{equation}\label{e6}
L_{cls}^f = E_{x,c}[-\log D_{cls}(c|G(x,c))]
\end{equation}
for fake shapes to optimize the generator. In practice, we employ cross-entropy loss for discrete expression or gender parts and mean-square loss for the continuous age part of the domain label. 

\textbf{The Cycle loss} employs a cycle consistency
loss~\cite{zhu2017unpaired} to preserve the domain-unrelated part while changing only the domain-related part in the input image, as
\begin{equation}\label{e7}
L_{cyc} = E_{x,c,c'}	\left \| x-G(G(x,c),c') \right \|_1.
\end{equation}

\textbf{The Reconstruction loss} makes the generator stable if the target domain label remains unchanged with respect to the source domain, as
\begin{equation}\label{e8}
L_{rec} = E_{x,c'}	\left \| x-G(x,c') \right \|_1.
\end{equation}

\textbf{The Symmetry loss} enforces the output facial shapes to be symmetric with respect to the central axis, as
\begin{equation}\label{e9}
L_{sym} = E_{y}	\left \| y-F(y) \right \|_1,
\end{equation}
where $F$ is a flip operator that flips horizontally for $y$ and then reverses the sign for the first channel. \textbf{We exclude the symmetric loss for asymmetric expressions in training.}  

\par \textbf{The full objective} functions to optimize $G$ and $D$ are weighted combinations of the above loss terms as 
\begin{equation}\label{e10}
L_D = -L_{adv} + \lambda_{cls} L^r_{cls}
\end{equation}
and
\begin{equation}\label{e11}
L_G = L_{adv} + \lambda_{cls}L^f_{cls} + \lambda_{cyc}L_{cyc} + \lambda_{rec}L_{rec} + \lambda_{sym}L_{sym},
\end{equation}
respectively.

\section{Experiments}

In this section, we carry out experiments on a public dataset~\cite{yang2020facescape} to demonstrate the effectiveness of the proposed method\footnote{We use the FaceScape dataset to train our model because it is the only one so far that provides registered 3D shapes with various attributes.}. We also show that the trained model is generalizable to other datasets~\cite{cudeiro2019capture,wang2022faceverse} combined with proper registrations.

\subsection{Dataset \& Labels}
The FaceScape dataset~\cite{yang2020facescape} provides a set of topological uniform (registered) samples with the NICP~\cite{amberg2007optimal} method. These samples cover $847$ identities, $20$ expressions, age ranges from $16$ to $70$, and $2$ different genders. We select $14,486$ out of all $18,760$ samples to exclude some noisy registered results. Among the selected samples, we further choose $10,005$ samples for training, $32$ samples as a mini-batch for validation using the \emph{leave-one-out principle}~\cite{hastie2009elements}, and the rest samples for testing. The identities of the training and testing samples are disjoint. 

\begin{figure}[htb]
	\begin{center}
		\includegraphics[width=1\linewidth]{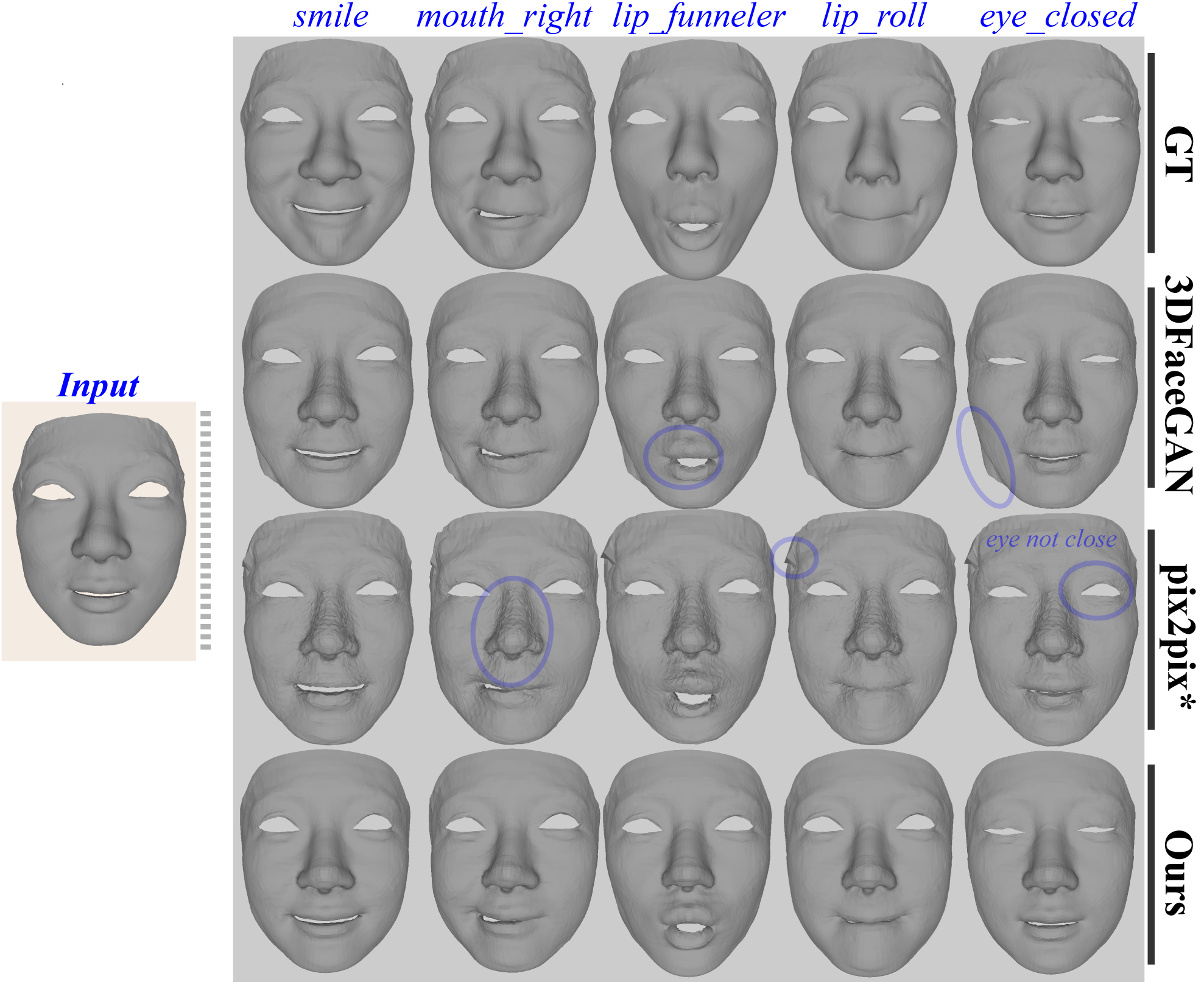}
	\end{center}
	\caption{Qualitative comparison for an exemplar input face. ``GT'' denotes the ground truth. Some artifacts and failure cases by the two baseline methods are marked with circles.}
	\label{fig_comparison}
\end{figure} 

One advantage of the proposed method is that it learns multi-domain translation with a unified GAN. To this end, we customize a domain label vector $c=\{c_i|i=1,2,...23\}$ of length $23$. The first $20$ dimensions and the following $21-22$ dimensions are \textbf{one-hot binary} expression and gender label, respectively. And the last dimension is a normalized \textbf{one-hot number} ranged by $[-1,1]$ for the ages.  

\subsection{Training Details \& Hyper-parameter Selection}
We implement all the networks with the Pytorch platform. The generator and discriminator are alternated \emph{one-by-one} in the training process. Adam~\cite{Diederik2015Adam} optimizer is used. The total iterations (mini-batches) are set to $800,000$. The first $400,000$ iterations adopt a fixed learning rate of $1e-4$. In the last $400,000$ iterations, the learning rate decreases to $1e-6$ linearly for every $2,000$ iterations. It takes about $50$ hours on a single GPU (specified as NVIDIA RTX 3090) to train the proposed networks. In the training process, we augment the data by random scaling and rotation by a range of $[0.9,1.1]$ and $[-10',10']$ for $3$ Euler angles, respectively.

The hyper-parameter settings for the weights of different loss functions in Eq.~\ref{e4} and Eq.~\ref{e11} follow a \emph{trial-and-error} principle guided by the visualized results of the validation set. We also follow some general settings in existing works~\cite{arjovsky2017wasserstein,choi2018stargan} and adjust the parameters until the loss curves keep oscillating, as a common way for tuning GAN. In our experiment, we set $\lambda_{cls}^C=0.02$, $\lambda_{cls}^M=0.05$, $\lambda_{cyc}=2$, $\lambda_{rec}=0.1$, $\lambda_{sym}=0.5$, and $\lambda_{gp}=0.2$, where the superscripts $C$ and $M$ on $\lambda_{cls}$ denote binary cross-entropy loss for classification (for expression and gender labels) and mean-square error loss for regression (for age label), respectively.

\subsection{Comparison for Expression Translation}

Previous works on deep learning based attribute translation mostly focus on 2D images~\cite{choi2018stargan} or 3D texture~\cite{wu2021adversarial}. There is seldom work on 3D facial shape attribute translation. However, there are some traceable works on 3D face generation with expressions. Therefore, we implement two baseline models which are close to ours to the best of our knowledge for comparisons. Since the UV map in~\cite{Stylianos20203DFaceGAN} cannot be reproduced exactly, we use our proposed geometric map instead.

\begin{itemize}
	\item \textbf{3DFaceGAN}~\cite{Stylianos20203DFaceGAN} is the first GAN tailored towards modeling the distribution of 3D facial surfaces. It can be used for 3D facial expression translation, which employs Multivariate-Gaussian decomposition and supervision to handle expressions on a UV map. 
	
	\item \textbf{Pix2pix}~\cite{isola2017image} is a widely used GAN for image translation applications. We blend our proposed geometric map with the official implementation~\cite{isola2017image} for pix2pix. We also constructed paired 3D shape data for neutral and other expressions for training. 
\end{itemize}

\begin{table}[htb]   
	\centering 
	\small 
	\begin{tabular}{lcc}
		\toprule
		Method & MSE-V (mm) &  MSE-N (degree)\\   
		\midrule 
		3DFaceGAN~\cite{Stylianos20203DFaceGAN}  & $1.05$  & $0.99$ \\ 
		Pix2pix*~\cite{isola2017image}  & $2.33$ & $3.47$ \\
		Our work  & $\bf{0.84}$ & $\bf{0.51}$ \\   
		\bottomrule
	\end{tabular}
	\caption{Comparisons of MSE on vertices and normal. The mark * denotes a combination with the proposed geometric map.}
	\label{Table_MSE}
\end{table}

We select neutral expressions for all subjects and transfer it to other expressions in the test set. After that, we compare the per-vertex and corresponding normal mean square error (\textbf{MSE-V} and \textbf{MSE-N}) between the generated results and the ground-truth ones after rigid registration of each pair of them by iterative closest point (\textbf{ICP})~\cite{paul1992registration} method. Table~\ref{Table_MSE} and Figure~\ref{fig_comparison} show the quantitative and qualitative comparisons. We can see that our proposed method performs over the pix2pix method by a large margin even without the paired setting. This is mainly due to the effective usage of data from multiple domains. In contrast, the pix2pix model only makes use of limited data pairs from two domains. Although the improvement of our results on MSE over 3DFaceGAN is not very salient, our qualitative results are visually better, especially on the edges of the mesh. We owe our success to the multi-domain translation framework and proper loss functions. Note that a result that is different from the ground truth is reasonable, since our method is supervised only by expression labels rather than pixel-wise errors. 

\subsection{Multi-domain Translation}

\begin{figure}[htb]
	\begin{center}
		\includegraphics[width=1\linewidth]{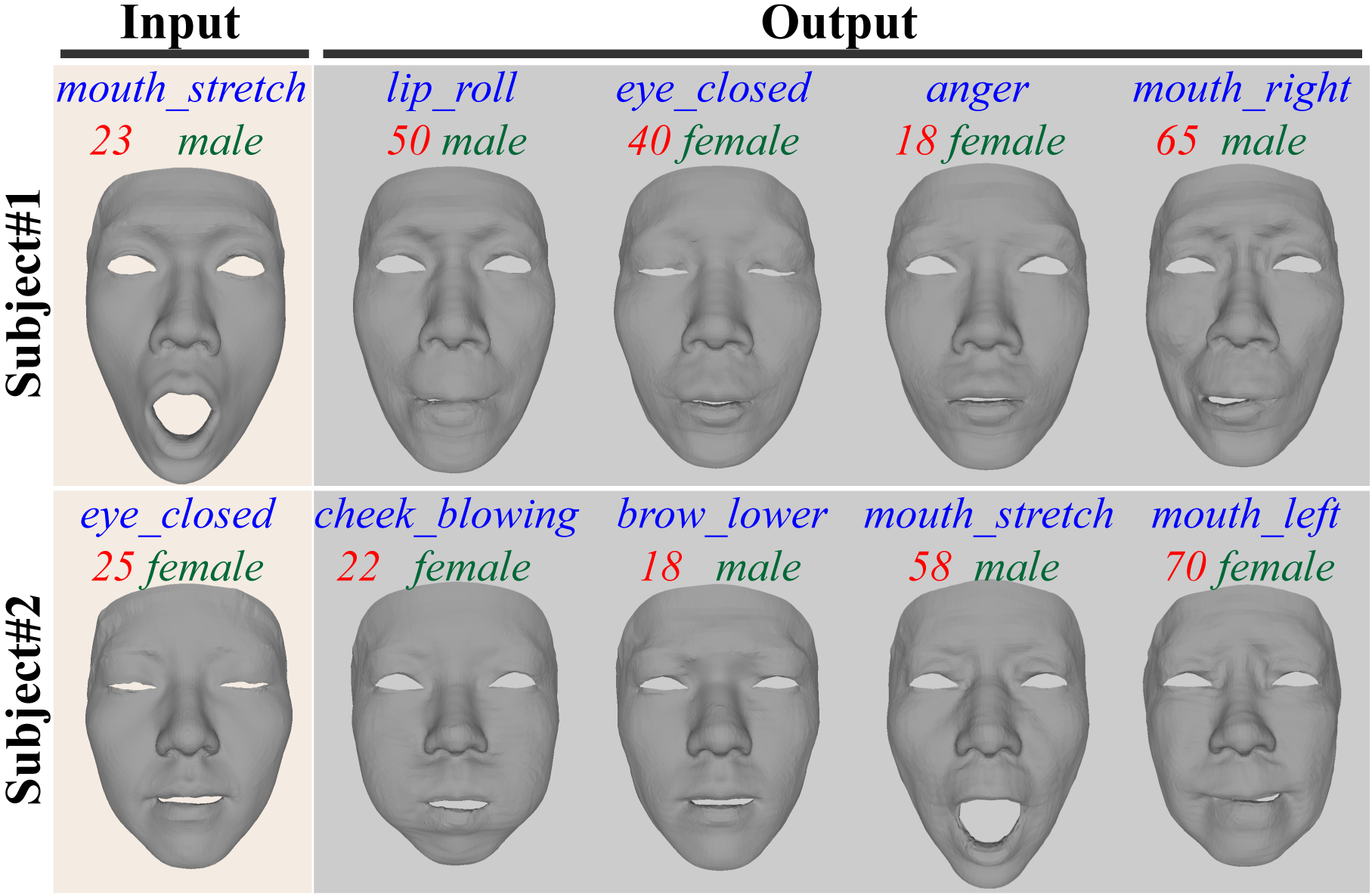}
	\end{center}
	\caption{Two qualitative examples for outputs with various attributes which are mostly different from the inputs. }
	\label{fig_multi_domain}
\end{figure}

In addition to expression transfer, our proposed framework is able to manipulate age and gender continuously\footnote{Please refer to Figure~\ref{fig_show} and more qualitative results on the supplementary material.}, which is a notable advantage over the existing works. Specifically, we only train a single network for expression, age, and gender translation tasks. Furthermore, the proposed method also supports input in various domains that are not limited to neutral such that it can broaden the applications to produce user-defined attributes~(see Figure~\ref{fig_multi_domain}). 

\begin{figure*}[htb]
	\begin{center}
		\includegraphics[width=1\linewidth]{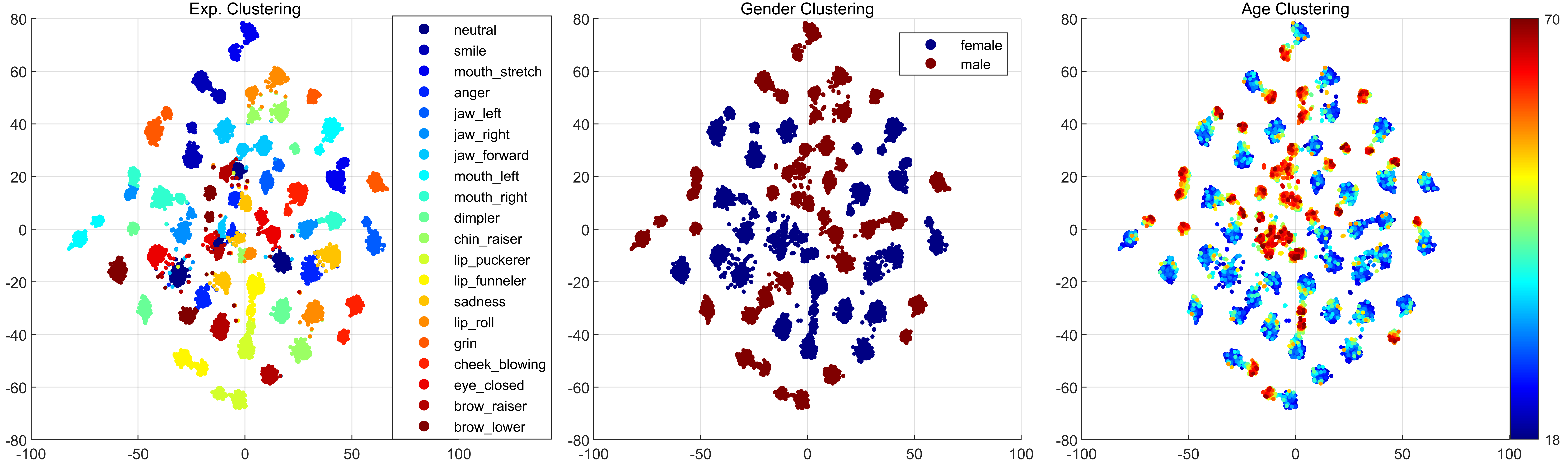}
	\end{center}
	\caption{Clustering of generated data with the t-SNE~\cite{van2008visualizing} method for different expressions, genders, and ages, respectively.}
	\label{fig_clustering}
\end{figure*} 

\subsection{Clustering Results for Feature Representation}
In this experiment, we project the output feature vectors at the classification head of the discriminator network onto a latent 2D plane to view the clustering effect of
data. We employ a t-distributed stochastic neighbor embedding (t-SNE)~\cite{van2008visualizing} method for latent space visualization. Figure~\ref{fig_clustering} shows the results marked with expression, age, and gender labels for some randomly generated samples from the testing samples in the FaceScape dataset. We can see that different classes are well separated and show noteworthy clustering effects, which implies that the trained model learns the representation of data of different domains effectively. Note that the three sub-figures for different attributes share the same representations. 

\begin{figure}[htb]
	\begin{center}
		\includegraphics[width=1\linewidth]{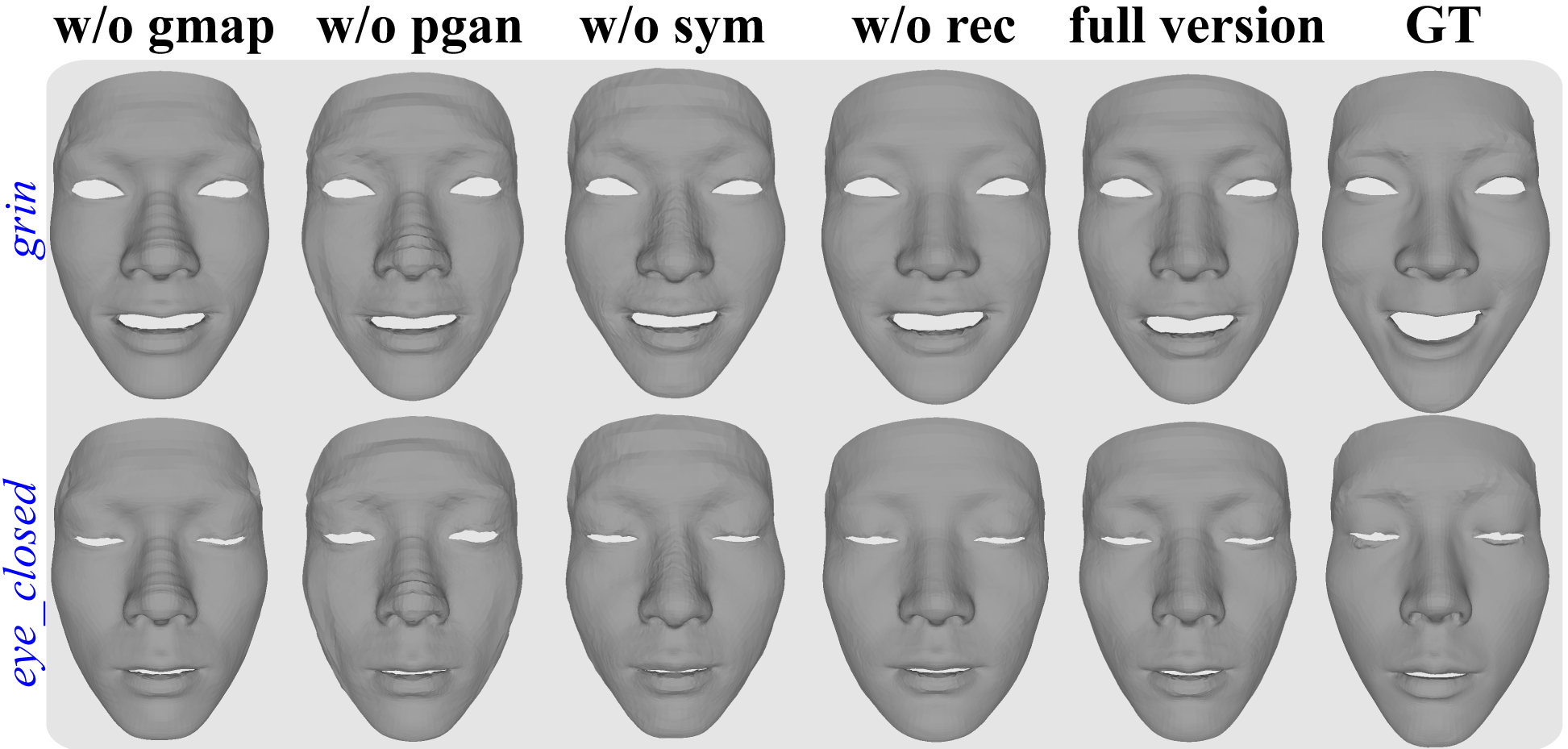}
	\end{center}
	\caption{Qualitative results of $2$ generated expressions for an exemplar face. The ground truths are also included for reference.}
	\label{fig_ablation}
\end{figure}

\subsection{Ablation Studies}

We now conduct experiments to learn the effects of different components of the proposed framework. The generator architectures is borrowed from the starGANs~\cite{choi2018stargan,choi2020stargan}. Some other components that are new to the existing works include: 
\begin{itemize}
	\item The square and symmetric geometric map. We replace it with the plain harmonic UV-map instead for the ablation study. This considers that the plain UV map is used in many existing works~\cite{wu2021adversarial, Stylianos20203DFaceGAN}.
	\item The Pyramid-GAN architecture. We replace the Pyramid-GAN architecture with Patch-GAN~\cite{li2016precomputed,choi2018stargan} only in the second last feature layer for ablation study. The Pyramid-GAN is in fact a generalization of Patch-GAN to diverse feature expansions.
	\item The symmetric loss function. We neglect this loss function for the ablation study.
	\item The reconstruction loss function. We neglect this loss function for the ablation study.
\end{itemize}

\begin{table}[htb]   
	\centering  
	\small
	\begin{tabular}{lcc}
		\toprule 
		Method & MSE-V (mm) &  MSE-N (degree)\\   
		\midrule 
		W/O geometric map & $1.31$  & $1.17$ \\ 
		W/O pyramid GAN  & $1.06$ & $1.13$ \\
		W/O symmetric loss  & $0.91$ & $0.78$ \\   
		W/O reconstruction loss  & $2.79$ & $1.98$ \\ 
		Full version  & $\bf{0.84}$ & $\bf{0.51}$ \\  
		\bottomrule
	\end{tabular}
	\caption{Comparisons on MSE for the ablation study.}
	\label{Table_MSE_Ablation}
\end{table}

\begin{figure}[htb]
	\begin{center}
		\includegraphics[width=1\linewidth]{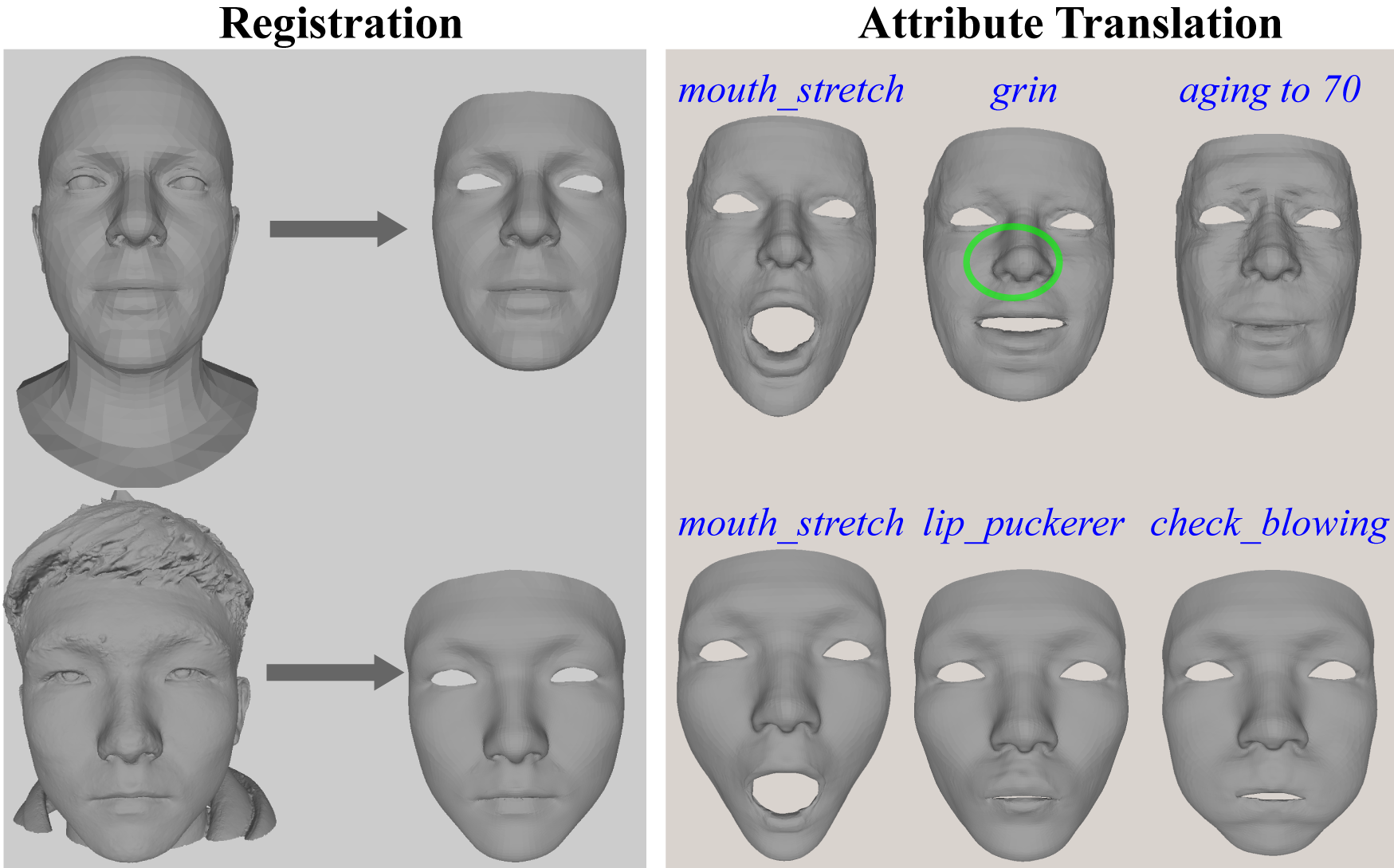}
	\end{center}
	\caption{Generalization to two exemplar faces from VOCA~\cite{cudeiro2019capture} and FaceVerse~\cite{wang2022faceverse} datasets, respectively.}
	\label{fig_generalize}
\end{figure}

We train the networks with the above settings, respectively. Table~\ref{Table_MSE_Ablation} shows the quantitative results in terms of the aforementioned MSE scores. We can see that each element contributes to some gains, among which the geometric map and the identity loss take effects most. Some qualitative results are also selected for a better understanding of the underlying mechanism of each proposed component, as in Figure~\ref{fig_ablation}. We observe that: 1) the geometric map contributes to realistic details since it handles a fundamental one-to-one 3D-to-2D mapping problem; 2) the Pyramid-GAN architecture reduces some artifacts over Patch-GAN; 3) the symmetric loss utilizes some prior knowledge of face for more stable and noiseless results; 4) the reconstruction loss resists identity drift. It is worth noting that the identity drift is not common in image-oriented tasks, however, it is a problem in the shape-oriented task in this work. This is because the pixel-wise values rather than 2D patterns are correlated to 3D shapes directly. 

\subsection{Generalization Test \& Limitations}

The deep learning methods on image data often suffer from generalization problems. However, in this work, we find that the trained model on FaceScape is well generalizable to other datasets if the target face is registered to a common template properly. We register the template to some samples from two other datasets (VOCA~\cite{cudeiro2019capture} and FaceVerse~\cite{wang2022faceverse}), and then feed the registered samples into the trained model. Figure~\ref{fig_generalize} shows two examples for attribute translation. Overall, 
both of the two samples are translated to the target attributes well. The reason should be that facial shape data are much simpler than texture data, since 3D shapes are immune to illumination and pose changes. However, some unexpected artifacts appear on the nose region  (marked with green circles ) in the top example. This is due to the intrinsic bias of the training set, \textit{i.e.}, all training samples are east Asian while the testing example is Caucasian. In the future, we will try to train a model with richer data that cover various ethnicities for better generalization. 

\section{Conclusion}   

In this paper, we propose an unpaired end-to-end adversarial learning framework for multi-domain 3D facial shape attribute translation.  Given an input 3D facial shape, the proposed framework is capable of synthesizing realistic 3D facial shapes with various expressions, genders, and ages with a unified generator network. The key element of our proposed framework is the canonical representation
of 3D faces by a square and symmetric geometric map, enabling effective learning on facial surfaces. Others include a Pyramid-GAN architecture and task-related loss functions, enabling unified and unpaired training with 3D data on the geometric map robustly. Extensive experiments demonstrate the effectiveness of the proposed method for the translation of various facial attributes. We hope this work will be helpful for future research and applications.

\section{Supplementary Material}

\subsection{Details for Geometric Mapping}

\begin{figure}[htb]
	\begin{center}
		\includegraphics[width=1\linewidth]{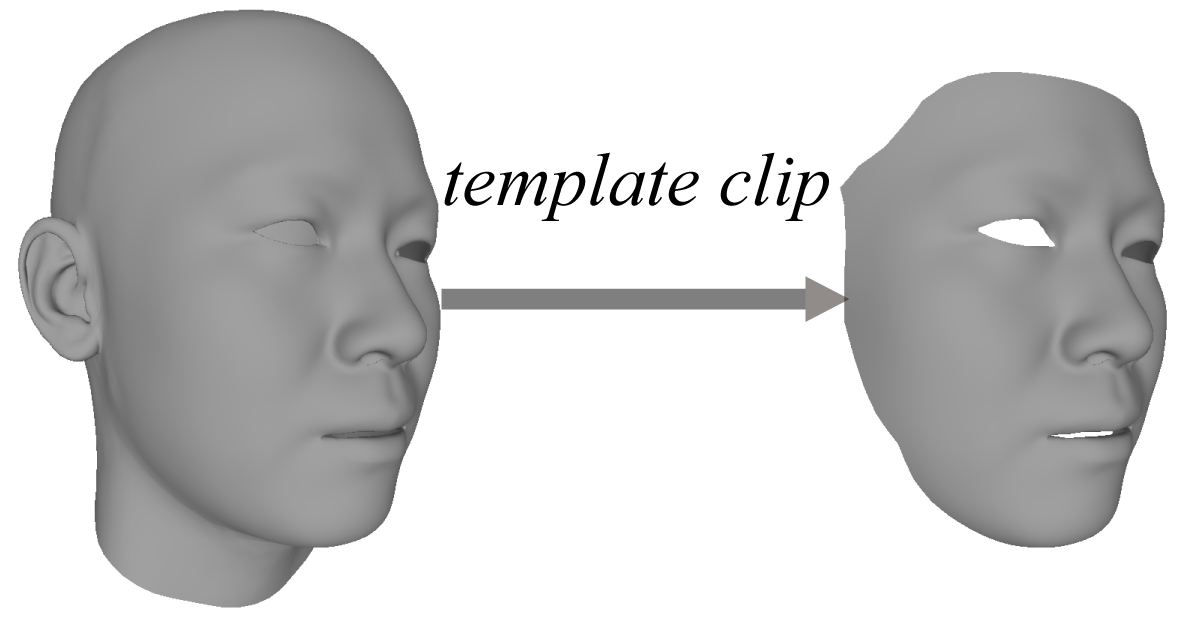}
	\end{center}
	\caption{Illustration for the clipped template.} 
	\label{fig_clipped}
\end{figure}

We clip the frontal part (see Figure~\ref{fig_clipped}) of the original 3D template on the FaceScape dataset~\cite{yang2020facescape}, considering that the variations of facial attributes (expression, age, and gender) are manifest only in the frontal parts in current 3D scans. The resulting resolution of the clipped face is $10,857$ vertices. The size of the geometric map (\textbf{Gmap}) is designed to be $128\times128\times3$, being a trade-off of computational efficiency and representation accuracy. The resolution for the Gmap ($128\times128=16384$) is also on par with the resolution of the clipped template, thereby being sufficient to represent the details of 3D facial shapes.

\begin{figure}[htb]
	\begin{center}
		\includegraphics[width=1\linewidth]{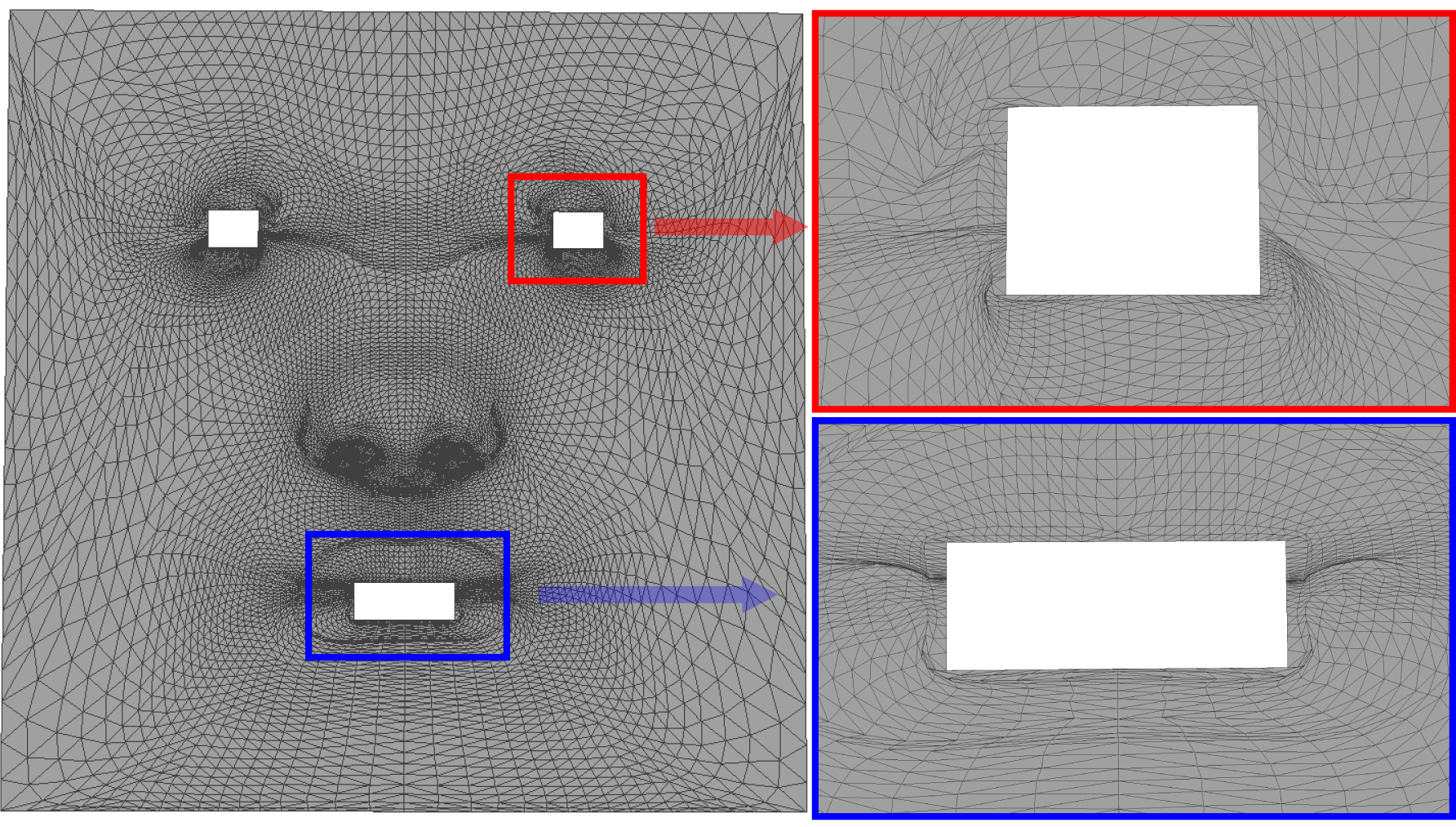}
	\end{center}
	\caption{The visualized details of the geometric map. Please zoom in to view the structures.} 
	\label{fig_details}
\end{figure}

In addition, the details of the Gmap are shown in Figure~\ref{fig_details}. It preserves the adjacency relationship of all vertices on the original 3D mesh in a \textbf{local least-square} sense while being \textbf{square and symmetric}. It also makes the mapping from 3D mesh to 2D geometric map to be one-to-one for each vertex, avoiding triangle flipping which is correlated to interpolation errors. We can see there is seldom triangle flipping even in the most difficult regions, \emph{i.e.} inner mouth and eye surroundings. 

\begin{figure}[htbp]
	\begin{center}
		\includegraphics[width=1\linewidth]{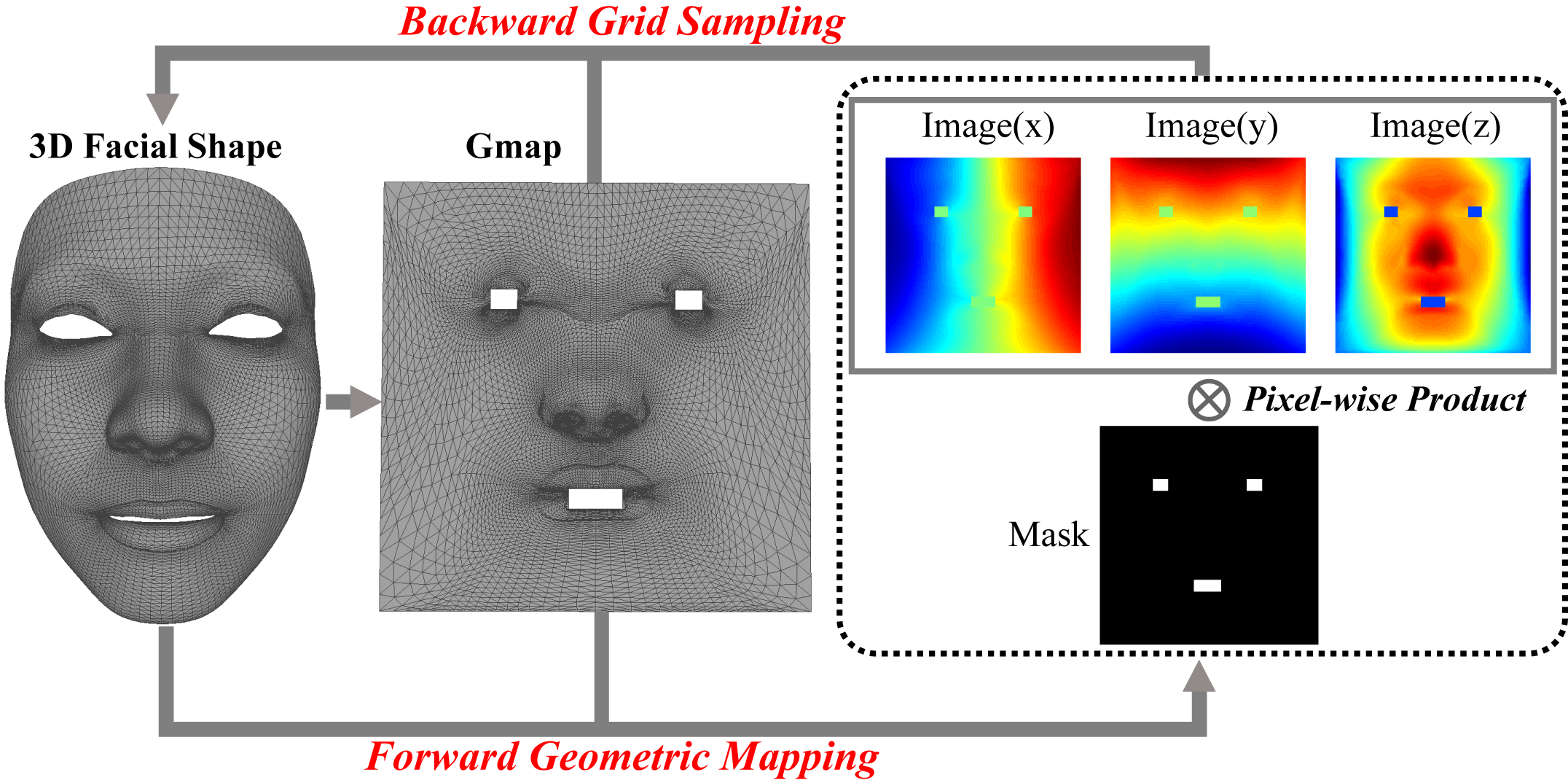}
	\end{center}
	\caption{Sampling the locations between a 3D facial shape and its representation on the Gmap.}
	\label{fig_gmap}
\end{figure} 

\begin{table*}[htb]
	\footnotesize
	\begin{center}
		\begin{tabular}{ccccccc}
			\hline
			\textbf{Index} & \textbf{Type} & \textbf{Kernel} & \textbf{Stride} & \textbf{Output} & \textbf{Others} & \textbf{Appended Loss}\\
			\hline
			1&Input shape& - &  - & $3 \times 10857$ & -& -\\
			\hline
			2&Geometric mapping & - &  - & $3 \times 128 \times 128$ & Mask Area set to $0$& -\\
			\hline
			3&Label cat. & - &  - & $26 \times 128 \times 128$ & -& -\\
			\hline
			\multirow{3}{*}{4} & Conv. & $7\times 7$ & $1\times 1$ & $64 \times 128 \times 128$ & IN+ReLU& -\\
			&Conv. & $4\times 4$ & $2\times 2$ & $128 \times 64 \times 64$ & IN+ReLU& -\\
			&Conv. & $4\times 4$ & $2\times 2$ & $256 \times 32 \times 32$ & IN+ReLU& -\\
			\hline
			5 &Residual blocks $\times 6$ & - & - & $256 \times 32 \times 32$ & - & -\\
			\hline
			\multirow{2}{*}{6} &DeConv. & $4\times4$ & $2\times 2$ & $128 \times 64 \times 64$ & IN+ReLU & -\\
			& DeConv. & $4\times4$ & $2\times 2$ & $64 \times 128 \times 128$ & IN+ReLU & -\\
			\hline
			7&Conv. & $7\times7$ & $1\times 1$ & $3 \times 128 \times 128$ & Mask Area set to $0$ & Symmetric loss\\
			\hline
			8&Bilinear grid sampling & - & - & $3 \times 10857$ & Mask & Cycle\&Reconstruction loss\\
			\hline
		\end{tabular}
	\end{center}
	\caption{The architecture of the generator network. The padding size of each layer is determined to be compatible with the input and output feature sizes. ``IN'' denotes for instance normalization operation. ``ReLU'' denotes for rectified linear unit activation.}
	\label{table:generator}
\end{table*}

\begin{table*}[htb]
	\footnotesize
	\begin{center}
		\begin{tabular}{ccccccc}
			\hline
			\textbf{Index} & \textbf{Type} & \textbf{Kernel} & \textbf{Stride} & \textbf{Output} & \textbf{Others} & \textbf{Appended Loss}\\
			\hline
			1&Input shape& - &  - & $3 \times 10857$ & -& -\\
			\hline
			2&Geometric mapping & - &  - & $3 \times 128 \times 128$ & Mask Area set to $0$& -\\
			\hline
			\multirow{6}{*}{3} & Conv. & $4\times 4$ & $2\times 2$ & $64 \times 64 \times 64$ & LReLU& -\\
			&Conv. & $4\times 4$ & $2\times 2$ & $128 \times 32 \times 32$ & LReLU& Conv. + Cat. + Adv. loss\\
			&Conv. & $4\times 4$ & $2\times 2$ & $256 \times 16 \times 16$ & LReLU& -\\
			&Conv. & $4\times 4$ & $2\times 2$ & $512 \times 8 \times 8$ & LReLU& Conv. + Cat. + Adv. loss\\
			&Conv. & $4\times 4$ & $2\times 2$ & $1024 \times 4 \times 4$ & LReLU& -\\
			&Conv. & $4\times 4$ & $2\times 2$ & $2048 \times 2 \times 2$ & LReLU& Conv. + Cat. + Adv. loss\\
			\hline
			4&Conv.\&Output for class & $2\times2$ & $1\times 1$ & $23 \times 1 \times 1$ & - & Classification loss\\
			\hline
		\end{tabular}
	\end{center}
	\caption{The architecture of the discriminator network. The padding size of each layer is determined to be compatible with the input and output feature sizes. ``LReLU'' denotes leaky ReLU activation. The Conv. operations listed in the appended loss denote convolutions to feature size $1$ by $3\times3$ kernels and $1\times1$ strides. The output features for the $3$ pyramid layers are then flattened and concatenated to fed into the adversarial loss.}
	\label{table:discriminator}
\end{table*}

Sampling the locations between a 3D shape and its representation on the 2D Gmap (on an image grid) involves bi-directional mappings between the 3D shape $\mathcal V$ and its vertex locations on the 2D image grid. The mappings are as follows (also refer to Figure~\ref{fig_gmap}).
\begin{itemize}
	\item \noindent \textbf{Forward mapping} is computed by \emph{barycentric interpolation}. Suppose a pixel $P=\{x,y\}$ on the image grid lies inside a triangle $\Delta P_1P_2P_3$ indexed by $i_1i_2i_3$ on the geometric map, then the shape coding $I_{x,y}$ is
	\begin{equation}\label{e1}
	I_{x,y}=w_1v_{i_1}+w_2v_{i_2}+w_3v_{i_3} (v_{i_1},v_{i_2},v_{i_3} \in \mathcal V),
	\end{equation}
	where 
	\begin{equation}\label{e2}
	\begin{aligned}
	w_1&=({\overrightarrow{PP_2}\times \overrightarrow{PP_3}})/({\overrightarrow{P_1P_2}\times \overrightarrow{P_1P_3}}),\\
	w_2&=({\overrightarrow{PP_3}\times \overrightarrow{PP_1}})/({\overrightarrow{P_1P_2}\times \overrightarrow{P_1P_3}}),\\
	w_3&=({\overrightarrow{PP_1}\times \overrightarrow{PP_2}})/({\overrightarrow{P_1P_2}\times \overrightarrow{P_1P_3}}),
	\end{aligned}
	\end{equation}
	and $\times$ is the outer-product between two 2D vectors.
	
	\item \noindent \textbf{Backward mapping} is computed by \emph{bilinear grid sampling} from the image grid to the geometric map. 
\end{itemize}

\subsection{Detailed Network Architecture}   

\begin{figure*}[htb]
	\begin{center}
		\includegraphics[width=1\linewidth]{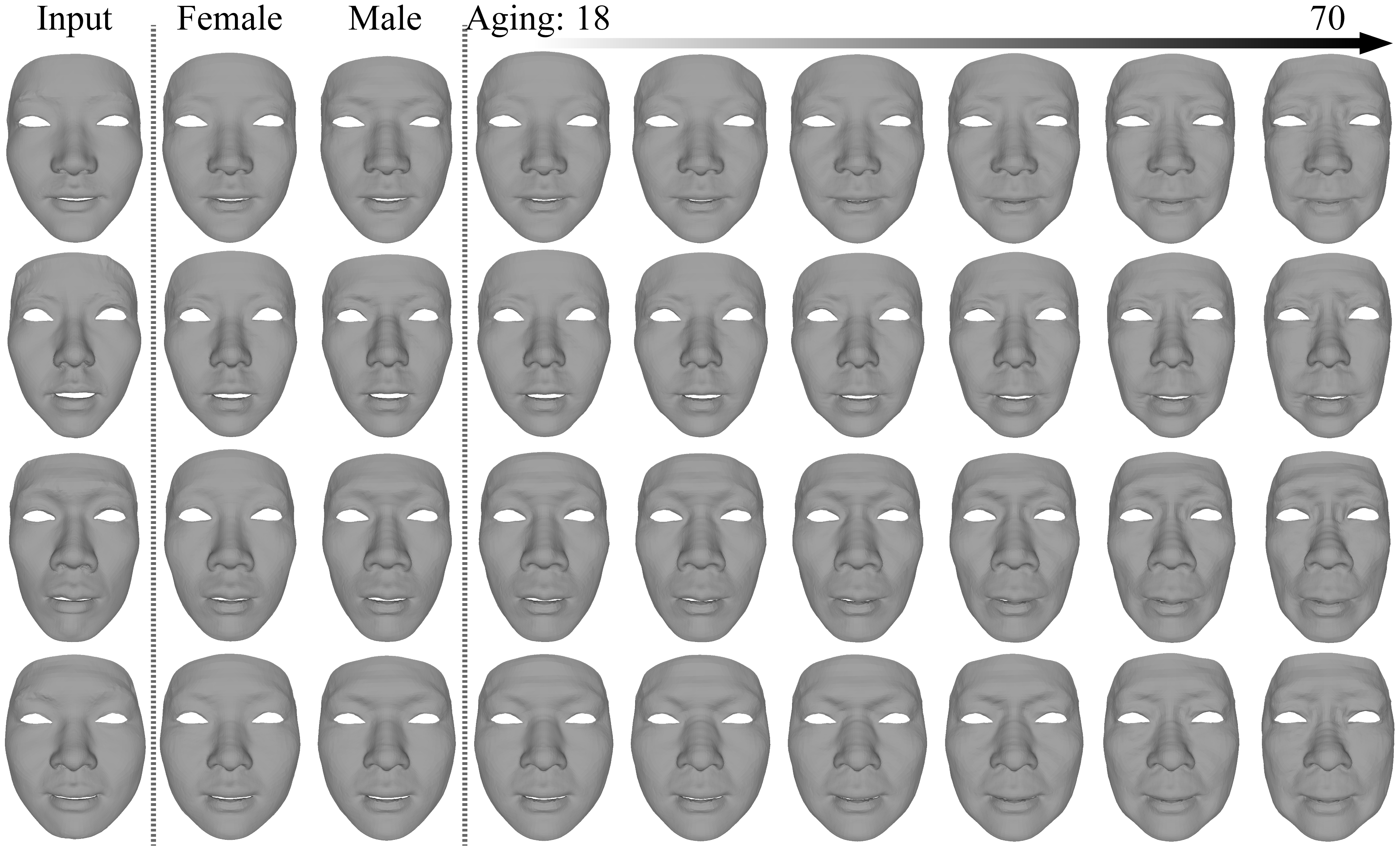}
	\end{center}
	\caption{Some additional results for gender and age translations.} 
	\label{fig_gender_age_sup}
\end{figure*} 

The detailed architectures for the generator and the discriminator of the proposed adversarial learning framework are elaborated in Table~\ref{table:generator} and Table~\ref{table:discriminator} for ease of reproductively, respectively. \textbf{We will also release our code.}

\subsection{Additional Qualitative Results}   

We have mentioned in the main manuscript that our method is also capable of translating gender and age. Figure~\ref{fig_gender_age_sup} shows additional results for some authorized samples in the test set of FaceScape~\cite{yang2020facescape}. In addition, the propose method supports both continous (\textit{e.g.} the expressions and genders) and discrete attribute labels (\textit{e.g.} the ages). We also suggest that the fractional labels for expressions can be acquired by
linear interpolations on the output directly. Figure~\ref{fig_exp} shows continous variations from neural to certain expressions. Therefore, our proposed method is capable of generating realistic shapes with different attributes given an input 3D facial shape. 
\begin{figure*}
	\begin{center}
		\subfigure{\animategraphics[width=0.4\linewidth,autoplay=True,palindrome,controls]{24}{gif/}{0}{10}}	
		\subfigure{\animategraphics[width=0.4\linewidth,autoplay=True,palindrome,controls]{24}{gif2/}{0}{10}}
	\end{center}
	\caption{Two examples for continous variations of the expressions.} 
	\label{fig_exp}
\end{figure*}

{\small
\bibliographystyle{ieee_fullname}
\bibliography{egbib}

\begin{thebibliography}{10}\itemsep=-1pt

\bibitem{abdal2021styleflow}
Rameen Abdal, Peihao Zhu, Niloy~J Mitra, and Peter Wonka.
\newblock Styleflow: Attribute-conditioned exploration of stylegan-generated
  images using conditional continuous normalizing flows.
\newblock {\em ACM Transactions on Graphics}, 40(3):1--21, 2021.

\bibitem{abrevaya2019decoupled}
Victoria~Fern{\'a}ndez Abrevaya, Adnane Boukhayma, Stefanie Wuhrer, and Edmond
  Boyer.
\newblock A decoupled 3d facial shape model by adversarial training.
\newblock In {\em IEEE International Conference on Computer Vision}, pages
  9419--9428, 2019.

\bibitem{amberg2009weight}
Brian Amberg, Pascal Paysan, and Thomas Vetter.
\newblock Weight, sex, and facial expressions: On the manipulation of
  attributes in generative 3d face models.
\newblock In {\em International Symposium on Advances in Visual Computing},
  pages 875--885, 2009.

\bibitem{amberg2007optimal}
Brian Amberg, Sami Romdhani, and Thomas Vetter.
\newblock Optimal step nonrigid icp algorithms for surface registration.
\newblock In {\em IEEE Conference on Computer Vision and Pattern Recognition},
  pages 1--8, 2007.

\bibitem{arjovsky2017wasserstein}
Martin Arjovsky, Soumith Chintala, and L{\'e}on Bottou.
\newblock Wasserstein generative adversarial networks.
\newblock In {\em International Conference on Machine Learning}, pages
  214--223, 2017.

\bibitem{bagautdinov2018modeling}
Timur Bagautdinov, Chenglei Wu, Jason Saragih, Pascal Fua, and Yaser Sheikh.
\newblock Modeling facial geometry using compositional vaes.
\newblock In {\em IEEE Conference on Computer Vision and Pattern Recognition},
  pages 3877--3886, 2018.

\bibitem{bao2021high}
Linchao Bao, Xiangkai Lin, Yajing Chen, Haoxian Zhang, Sheng Wang, Xuefei Zhe,
  Di Kang, Haozhi Huang, Xinwei Jiang, Jue Wang, et~al.
\newblock High-fidelity 3d digital human head creation from rgb-d selfies.
\newblock {\em ACM Transactions on Graphics}, 41(1):1--21, 2021.

\bibitem{paul1992registration}
Paul~J. Besl and Neil~D. McKay.
\newblock A method for registration of 3-d shapes.
\newblock {\em IEEE Transactions on Pattern Analysis and Machine Intelligence},
  14(2):239--256, 1992.

\bibitem{blanz1999morphable}
Volker Blanz and Thomas Vetter.
\newblock A morphable model for the synthesis of 3d faces.
\newblock In {\em Annual conference on Computer Graphics and Interactive
  Techniques}, pages 187--194, 1999.

\bibitem{booth2018large}
James Booth, Anastasios Roussos, Allan Ponniah, David Dunaway, and Stefanos
  Zafeiriou.
\newblock Large scale 3d morphable models.
\newblock {\em International Journal of Computer Vision}, 126(2):233--254,
  2018.

\bibitem{chang2021learning}
Jia-Ren Chang, Yong-Sheng Chen, and Wei-Chen Chiu.
\newblock Learning facial representations from the cycle-consistency of face.
\newblock In {\em IEEE International Conference on Computer Vision}, pages
  9680--9689, 2021.

\bibitem{cheng2019meshgan}
Shiyang Cheng, Michael Bronstein, Yuxiang Zhou, Irene Kotsia, Maja Pantic, and
  Stefanos Zafeiriou.
\newblock Meshgan: Non-linear 3d morphable models of faces.
\newblock {\em arXiv preprint arXiv:1903.10384}, 2019.

\bibitem{choi2018stargan}
Yunjey Choi, Minje Choi, Munyoung Kim, Jung-Woo Ha, Sunghun Kim, and Jaegul
  Choo.
\newblock Stargan: Unified generative adversarial networks for multi-domain
  image-to-image translation.
\newblock In {\em IEEE Conference on Computer Vision and Pattern Recognition},
  pages 8789--8797, 2018.

\bibitem{choi2020stargan}
Yunjey Choi, Youngjung Uh, Jaejun Yoo, and Jung-Woo Ha.
\newblock Stargan v2: Diverse image synthesis for multiple domains.
\newblock In {\em IEEE Conference on Computer Vision and Pattern Recognition},
  pages 8188--8197, 2020.

\bibitem{cudeiro2019capture}
Daniel Cudeiro, Timo Bolkart, Cassidy Laidlaw, Anurag Ranjan, and Michael~J
  Black.
\newblock Capture, learning, and synthesis of 3d speaking styles.
\newblock In {\em IEEE Conference on Computer Vision and Pattern Recognition},
  pages 10101--10111, 2019.

\bibitem{eck1995multiresolution}
Matthias Eck, Tony DeRose, Tom Duchamp, Hugues Hoppe, Michael Lounsbery, and
  Werner Stuetzle.
\newblock Multiresolution analysis of arbitrary meshes.
\newblock In {\em Annual Conference on Computer Graphics and Interactive
  Techniques}, pages 173--182, 1995.

\bibitem{egger2018occlusion}
Bernhard Egger, Sandro Sch{\"o}nborn, Andreas Schneider, Adam Kortylewski,
  Andreas Morel-Forster, Clemens Blumer, and Thomas Vetter.
\newblock Occlusion-aware 3d morphable models and an illumination prior for
  face image analysis.
\newblock {\em International Journal of Computer Vision}, 126(12):1269--1287,
  2018.

\bibitem{egger20203d}
Bernhard Egger, William~AP Smith, Ayush Tewari, Stefanie Wuhrer, Michael
  Zollhoefer, Thabo Beeler, Florian Bernard, Timo Bolkart, Adam Kortylewski,
  Sami Romdhani, et~al.
\newblock 3d morphable face models—past, present, and future.
\newblock {\em ACM Transactions on Graphics}, 39(5):1--38, 2020.

\bibitem{fan2019controllable}
Lijie Fan, Wenbing Huang, Chuang Gan, Junzhou Huang, and Boqing Gong.
\newblock Controllable image-to-video translation: A case study on facial
  expression generation.
\newblock In {\em AAAI Conference on Artificial Intelligence}, volume~33, pages
  3510--3517, 2019.

\bibitem{fan2023towards}
Zhenfeng Fan, Silong Peng, and Shihong Xia.
\newblock Towards fine-grained optimal 3d face dense registration: An iterative
  dividing and diffusing method.
\newblock {\em International Journal of Computer Vision}, pages 1--21, June
  2023.

\bibitem{gafni2021dynamic}
Guy Gafni, Justus Thies, Michael Zollhofer, and Matthias Nie{\ss}ner.
\newblock Dynamic neural radiance fields for monocular 4d facial avatar
  reconstruction.
\newblock In {\em Proceedings of the IEEE/CVF Conference on Computer Vision and
  Pattern Recognition}, pages 8649--8658, 2021.

\bibitem{gecer2020synthesizing}
Baris Gecer, Alexandros Lattas, Stylianos Ploumpis, Jiankang Deng, Athanasios
  Papaioannou, Stylianos Moschoglou, and Stefanos Zafeiriou.
\newblock Synthesizing coupled 3d face modalities by trunk-branch generative
  adversarial networks.
\newblock In {\em European Conference on Computer Vision}, pages 415--433,
  2020.

\bibitem{geng2018warp}
Jiahao Geng, Tianjia Shao, Youyi Zheng, Yanlin Weng, and Kun Zhou.
\newblock Warp-guided gans for single-photo facial animation.
\newblock {\em ACM Transactions on Graphics}, 37(6):1--12, 2018.

\bibitem{goodfellow2014generative}
Ian Goodfellow, Jean Pouget-Abadie, Mehdi Mirza, Bing Xu, David Warde-Farley,
  Sherjil Ozair, Aaron Courville, and Yoshua Bengio.
\newblock Generative adversarial nets.
\newblock {\em Advances in Neural Information Processing Systems}, 27, 2014.

\bibitem{grassal2022neural}
Philip-William Grassal, Malte Prinzler, Titus Leistner, Carsten Rother,
  Matthias Nie{\ss}ner, and Justus Thies.
\newblock Neural head avatars from monocular rgb videos.
\newblock In {\em Proceedings of the IEEE/CVF Conference on Computer Vision and
  Pattern Recognition}, pages 18653--18664, 2022.

\bibitem{gu2002geometry}
Xianfeng Gu, Steven~J Gortler, and Hugues Hoppe.
\newblock Geometry images.
\newblock In {\em Annual Conference on Computer Graphics and Interactive
  Techniques}, pages 355--361, 2002.

\bibitem{hastie2009elements}
Trevor Hastie, Robert Tibshirani, Jerome~H Friedman, and Jerome~H Friedman.
\newblock {\em The elements of statistical learning: data mining, inference,
  and prediction}, volume~2.
\newblock Springer, 2009.

\bibitem{he2016deep}
Kaiming He, Xiangyu Zhang, Shaoqing Ren, and Jian Sun.
\newblock Deep residual learning for image recognition.
\newblock In {\em IEEE Conference on Computer Vision and Pattern Recognition},
  pages 770--778, 2016.

\bibitem{he2019attgan}
Zhenliang He, Wangmeng Zuo, Meina Kan, Shiguang Shan, and Xilin Chen.
\newblock Attgan: Facial attribute editing by only changing what you want.
\newblock {\em IEEE Transactions on Image Processing}, 28(11):5464--5478, 2019.

\bibitem{huang2018multimodal}
Xun Huang, Ming-Yu Liu, Serge Belongie, and Jan Kautz.
\newblock Multimodal unsupervised image-to-image translation.
\newblock In {\em European Conference on Computer Vision}, pages 172--189,
  2018.

\bibitem{isobe2021multi}
Takashi Isobe, Xu Jia, Shuaijun Chen, Jianzhong He, Yongjie Shi, Jianzhuang
  Liu, Huchuan Lu, and Shengjin Wang.
\newblock Multi-target domain adaptation with collaborative consistency
  learning.
\newblock In {\em IEEE Conference on Computer Vision and Pattern Recognition},
  pages 8187--8196, 2021.

\bibitem{isola2017image}
Phillip Isola, Jun-Yan Zhu, Tinghui Zhou, and Alexei~A Efros.
\newblock Image-to-image translation with conditional adversarial networks.
\newblock In {\em IEEE conference on Computer Vision and Pattern Recognition},
  pages 1125--1134, 2017.

\bibitem{jackson2017large}
Aaron~S Jackson, Adrian Bulat, Vasileios Argyriou, and Georgios Tzimiropoulos.
\newblock Large pose 3d face reconstruction from a single image via direct
  volumetric cnn regression.
\newblock In {\em IEEE International Conference on Computer Vision}, pages
  1031--1039, 2017.

\bibitem{karras2019style}
Tero Karras, Samuli Laine, and Timo Aila.
\newblock A style-based generator architecture for generative adversarial
  networks.
\newblock In {\em IEEE Conference on Computer Vision and Pattern Recognition},
  pages 4401--4410, 2019.

\bibitem{KimKKL20}
Junho Kim, Minjae Kim, Hyeonwoo Kang, and Kwanghee Lee.
\newblock {U-GAT-IT:} unsupervised generative attentional networks with
  adaptive layer-instance normalization for image-to-image translation.
\newblock In {\em International Conference on Learning Representations}, 2020.

\bibitem{kim2017learning}
Taeksoo Kim, Moonsu Cha, Hyunsoo Kim, Jung~Kwon Lee, and Jiwon Kim.
\newblock Learning to discover cross-domain relations with generative
  adversarial networks.
\newblock In {\em International Conference on Machine Learning}, pages
  1857--1865, 2017.

\bibitem{Diederik2015Adam}
Diederik~P. Kingma and Jimmy Ba.
\newblock Adam: {A} method for stochastic optimization.
\newblock In Yoshua Bengio and Yann LeCun, editors, {\em International
  Conference on Learning Representations}, 2015.

\bibitem{ledig2017photo}
Christian Ledig, Lucas Theis, Ferenc Husz{\'a}r, Jose Caballero, Andrew
  Cunningham, Alejandro Acosta, Andrew Aitken, Alykhan Tejani, Johannes Totz,
  Zehan Wang, et~al.
\newblock Photo-realistic single image super-resolution using a generative
  adversarial network.
\newblock In {\em IEEE Conference on Computer Vision and Pattern Recognition},
  pages 4681--4690, 2017.

\bibitem{li2016precomputed}
Chuan Li and Michael Wand.
\newblock Precomputed real-time texture synthesis with markovian generative
  adversarial networks.
\newblock In {\em European Conference on Computer Vision}, pages 702--716,
  2016.

\bibitem{li2010example}
Hao Li, Thibaut Weise, and Mark Pauly.
\newblock Example-based facial rigging.
\newblock {\em ACM Transactions on Graphics}, 29(4):1--6, 2010.

\bibitem{lin2020towards}
Jiangke Lin, Yi Yuan, Tianjia Shao, and Kun Zhou.
\newblock Towards high-fidelity 3d face reconstruction from in-the-wild images
  using graph convolutional networks.
\newblock In {\em IEEE Conference on Computer Vision and Pattern Recognition},
  pages 5891--5900, 2020.

\bibitem{liu2018joint}
Feng Liu, Qijun Zhao, Xiaoming Liu, and Dan Zeng.
\newblock Joint face alignment and 3d face reconstruction with application to
  face recognition.
\newblock {\em IEEE Transactions on Pattern Analysis and Machine Intelligence},
  42(3):664--678, 2018.

\bibitem{liu2018disentangling}
Feng Liu, Ronghang Zhu, Dan Zeng, Qijun Zhao, and Xiaoming Liu.
\newblock Disentangling features in 3d face shapes for joint face
  reconstruction and recognition.
\newblock In {\em IEEE Conference on Computer Vision and Pattern Recognition},
  pages 5216--5225, 2018.

\bibitem{liu2017unsupervised}
Ming-Yu Liu, Thomas Breuel, and Jan Kautz.
\newblock Unsupervised image-to-image translation networks.
\newblock {\em Advances in Neural Information Processing Systems}, 30, 2017.

\bibitem{ma2021pixel}
Shugao Ma, Tomas Simon, Jason Saragih, Dawei Wang, Yuecheng Li, Fernando
  De~La~Torre, and Yaser Sheikh.
\newblock Pixel codec avatars.
\newblock In {\em Proceedings of the IEEE/CVF Conference on Computer Vision and
  Pattern Recognition}, pages 64--73, 2021.

\bibitem{mirza2014conditional}
Mehdi Mirza and Simon Osindero.
\newblock Conditional generative adversarial nets.
\newblock {\em arXiv preprint arXiv:1411.1784}, 2014.

\bibitem{Stylianos20203DFaceGAN}
Stylianos Moschoglou, Stylianos Ploumpis, Mihalis~A. Nicolaou, Athanasios
  Papaioannou, and Stefanos Zafeiriou.
\newblock 3dfacegan: Adversarial nets for 3d face representation, generation,
  and translation.
\newblock {\em International Journal of Computer Vision}, 128(10):2534--2551,
  2020.

\bibitem{odena2017conditional}
Augustus Odena, Christopher Olah, and Jonathon Shlens.
\newblock Conditional image synthesis with auxiliary classifier gans.
\newblock In {\em International Conference on Machine Learning}, pages
  2642--2651, 2017.

\bibitem{park2019deepsdf}
Jeong~Joon Park, Peter Florence, Julian Straub, Richard Newcombe, and Steven
  Lovegrove.
\newblock Deepsdf: Learning continuous signed distance functions for shape
  representation.
\newblock In {\em Proceedings of the IEEE/CVF conference on computer vision and
  pattern recognition}, pages 165--174, 2019.

\bibitem{ploumpis2020towards}
Stylianos Ploumpis, Evangelos Ververas, Eimear O'Sullivan, Stylianos
  Moschoglou, Haoyang Wang, Nick Pears, William~AP Smith, Baris Gecer, and
  Stefanos Zafeiriou.
\newblock Towards a complete 3d morphable model of the human head.
\newblock {\em IEEE Transactions on Pattern Analysis and Machine Intelligence},
  43(11):4142--4160, 2020.

\bibitem{pumarola2018ganimation}
Albert Pumarola, Antonio Agudo, Aleix~M Martinez, Alberto Sanfeliu, and
  Francesc Moreno-Noguer.
\newblock Ganimation: Anatomically-aware facial animation from a single image.
\newblock In {\em European conference on computer vision}, pages 818--833,
  2018.

\bibitem{ranjan2018generating}
Anurag Ranjan, Timo Bolkart, Soubhik Sanyal, and Michael~J Black.
\newblock Generating 3d faces using convolutional mesh autoencoders.
\newblock In {\em European Conference on Computer Vision}, pages 704--720,
  2018.

\bibitem{richard2021meshtalk}
Alexander Richard, Michael Zollh{\"o}fer, Yandong Wen, Fernando De~la Torre,
  and Yaser Sheikh.
\newblock Meshtalk: 3d face animation from speech using cross-modality
  disentanglement.
\newblock In {\em Proceedings of the IEEE/CVF International Conference on
  Computer Vision}, pages 1173--1182, 2021.

\bibitem{sheffer2007mesh}
Alla Sheffer, Emil Praun, Kenneth Rose, et~al.
\newblock Mesh parameterization methods and their applications.
\newblock {\em Foundations and Trends in Computer Graphics and Vision},
  2(2):105--171, 2007.

\bibitem{shen2020interfacegan}
Yujun Shen, Ceyuan Yang, Xiaoou Tang, and Bolei Zhou.
\newblock Interfacegan: Interpreting the disentangled face representation
  learned by gans.
\newblock {\em IEEE Transactions on Pattern Analysis and Machine Intelligence},
  44(4):2004--2018, 2020.

\bibitem{TangLXTS23}
Hao Tang, Hong Liu, Dan Xu, Philip H.~S. Torr, and Nicu Sebe.
\newblock Attentiongan: Unpaired image-to-image translation using
  attention-guided generative adversarial networks.
\newblock {\em IEEE Transactions on Neural Networks and Learning Systems},
  34(4):1972--1987, 2021.

\bibitem{tewari2021learning}
Ayush Tewari, Hans-Peter Seidel, Mohamed Elgharib, Christian Theobalt, et~al.
\newblock Learning complete 3d morphable face models from images and videos.
\newblock In {\em IEEE Conference on Computer Vision and Pattern Recognition},
  pages 3361--3371, 2021.

\bibitem{van2008visualizing}
Laurens Van~der Maaten and Geoffrey Hinton.
\newblock Visualizing data using t-sne.
\newblock {\em Journal of Machine Learning Research}, 9(11), 2008.

\bibitem{wang2022faceverse}
Lizhen Wang, Zhiyuan Chen, Tao Yu, Chenguang Ma, Liang Li, and Yebin Liu.
\newblock Faceverse: a fine-grained and detail-controllable 3d face morphable
  model from a hybrid dataset.
\newblock In {\em IEEE Conference on Computer Vision and Pattern Recognition},
  pages 20333--20342, 2022.

\bibitem{wu2021adversarial}
Yiqiang Wu, Ruxin Wang, Mingming Gong, Jun Cheng, Zhengtao Yu, and Dapeng Tao.
\newblock Adversarial uv-transformation texture estimation for 3d face aging.
\newblock {\em IEEE Transactions on Circuits and Systems for Video Technology},
  2021.

\bibitem{yang2018learning}
Hongyu Yang, Di Huang, Yunhong Wang, and Anil~K Jain.
\newblock Learning face age progression: A pyramid architecture of gans.
\newblock In {\em IEEE Conference on Computer Vision and Pattern Recognition},
  pages 31--39, 2018.

\bibitem{yang2020facescape}
Haotian Yang, Hao Zhu, Yanru Wang, Mingkai Huang, Qiu Shen, Ruigang Yang, and
  Xun Cao.
\newblock Facescape: a large-scale high quality 3d face dataset and detailed
  riggable 3d face prediction.
\newblock In {\em IEEE Conference on Computer Vision and Pattern Recognition},
  pages 601--610, 2020.

\bibitem{yang2015go}
Jiaolong Yang, Hongdong Li, Dylan Campbell, and Yunde Jia.
\newblock Go-icp: A globally optimal solution to 3d icp point-set registration.
\newblock {\em IEEE Transactions on Pattern Analysis and Machine Intelligence},
  38(11):2241--2254, 2015.

\bibitem{zhan2021unbalanced}
Fangneng Zhan, Yingchen Yu, Kaiwen Cui, Gongjie Zhang, Shijian Lu, Jianxiong
  Pan, Changgong Zhang, Feiying Ma, Xuansong Xie, and Chunyan Miao.
\newblock Unbalanced feature transport for exemplar-based image translation.
\newblock In {\em IEEE Conference on Computer Vision and Pattern Recognition},
  pages 15028--15038, 2021.

\bibitem{zhao20183d}
Jian Zhao, Lin Xiong, Jianshu Li, Junliang Xing, Shuicheng Yan, and Jiashi
  Feng.
\newblock 3d-aided dual-agent gans for unconstrained face recognition.
\newblock {\em IEEE Transactions on Pattern Analysis and Machine Intelligence},
  41(10):2380--2394, 2018.

\bibitem{zheng2021spatially}
Chuanxia Zheng, Tat-Jen Cham, and Jianfei Cai.
\newblock The spatially-correlative loss for various image translation tasks.
\newblock In {\em IEEE Conference on Computer Vision and Pattern Recognition},
  pages 16407--16417, 2021.

\bibitem{zheng2022avatar}
Yufeng Zheng, Victoria~Fern{\'a}ndez Abrevaya, Marcel~C B{\"u}hler, Xu Chen,
  Michael~J Black, and Otmar Hilliges.
\newblock Im avatar: Implicit morphable head avatars from videos.
\newblock In {\em Proceedings of the IEEE/CVF Conference on Computer Vision and
  Pattern Recognition}, pages 13545--13555, 2022.

\bibitem{zhu2017unpaired}
Jun-Yan Zhu, Taesung Park, Phillip Isola, and Alexei~A Efros.
\newblock Unpaired image-to-image translation using cycle-consistent
  adversarial networks.
\newblock In {\em IEEE International Conference on Computer Vision}, pages
  2223--2232, 2017.

\end{thebibliography}
}

\end{document}